\documentclass[draftclsnofoot,onecolumn,12pt]{IEEEtran}
\ifCLASSINFOpdf
\usepackage[pdftex]{graphicx}
\else
\usepackage[dvips]{graphicx}
\fi
\graphicspath{{figs/}}
\DeclareGraphicsExtensions{.pdf,.eps}
\usepackage{cite}
\usepackage{amsmath,amssymb,amsfonts}
\usepackage{graphicx}
\usepackage{textcomp}
\usepackage{subcaption}
\usepackage{cite}
\usepackage{color}
\usepackage[linesnumbered, ruled]{algorithm2e}
\usepackage{algpseudocode}
\usepackage{array}
\usepackage{url}
\usepackage{enumerate}
\usepackage{multirow}

\newtheorem{theorem}{Theorem}

\newtheorem{lemma}{Lemma}
\newtheorem{prop}{Proposition}
\newtheorem{defn}{Definition}
\begin{document}
\title{An Incentive Mechanism for Federated Learning in Wireless Cellular network: An Auction Approach}
%
%
% author names and IEEE memberships
% note positions of commas and nonbreaking spaces ( ~ ) LaTeX will not break
% a structure at a ~ so this keeps an author's name from being broken across
% two lines.
% use \thanks{} to gain access to the first footnote area
% a separate \thanks must be used for each paragraph as LaTeX2e's \thanks
% was not built to handle multiple paragraphs
%
\author{Tra Huong Thi Le,~\IEEEmembership{Student~Member,~IEEE,}\\
	Nguyen H.~Tran,~\IEEEmembership{Senior~Member,~IEEE,}
	Yan Kyaw Tun,
	Minh N.~H.~Nguyen,\\
	Zhu Han,~\IEEEmembership{Fellow~Member,~IEEE}
	and~Choong Seon Hong,~\IEEEmembership{Senior~Member,~IEEE}% <-this % stops a space
	\thanks{T.~H.~T.~Le, Yan Kyaw Tun, Minh N.~H.~Nguyen and C.S. Hong are with Dept. of Computer Science and Engineering, Kyung Hee University, Korea. e-mail: \{huong\_tra25, ykyawtun7, minhnhn, cshong\}@khu.ac.kr}% <-this % stops a space
	\thanks{N.~H.~Tran is with School of Computer Science, The University of Sydney, Sydney, NSW 2006, Australia. email: nguyen.tran@sydney.edu.au}% <-this % stops a space
	\thanks{Z.~Han is with Department of Electrical and Computer Engineering, University of Houston, Houston, TX 77004-4005 USA. email: zhan2@uh.edu}
	\thanks{Manuscript received XXX, XX, 2015; revised XXX, XX, 2015.}}

% The paper headers
\markboth{Journal of \LaTeX\ Class Files,~Vol.~14, No.~8, August~2015}%
{Shell \MakeLowercase{\textit{et al.}}: Bare Demo of IEEEtran.cls for IEEE Journals}

% If you want to put a publisher's ID mark on the page you can do it like
% this:
%\IEEEpubid{0000--0000/00\$00.00~\copyright~2015 IEEE}
% Remember, if you use this you must call \IEEEpubidadjcol in the second
% column for its text to clear the IEEEpubid mark.

% use for special paper notices
%\IEEEspecialpapernotice{(Invited Paper)}

% make the title area
\maketitle

% As a general rule, do not put math, special symbols or citations
% in the abstract or keywords.
\begin{abstract}
Federated Learning (FL) is a distributed learning framework that can deal with the distributed issue in machine learning and still guarantee high learning performance.
However, it is impractical that all users will sacrifice their resources to join the FL algorithm.
This motivates us to study the incentive mechanism design for FL. 
In this paper, we consider a FL system that involves one base station (BS) and multiple mobile users. The mobile users use their own data to train the local machine learning model, and then send the trained models to the BS, which generates the initial model, collects local models and constructs the global model.  
Then, we formulate the incentive mechanism between the BS and mobile users as an auction game where the BS is an auctioneer and the mobile users are the sellers.
In the proposed game, each mobile user submits its bids according to the minimal energy cost that the mobile users experiences in participating in FL.
To decide winners in the auction and maximize social welfare, we propose the primal-dual greedy auction mechanism.
The proposed mechanism can guarantee three economic properties, namely, truthfulness, individual rationality and efficiency. 
Finally, numerical results are shown to demonstrate the performance effectiveness of our proposed mechanism.
\end{abstract}

% Note that keywords are not normally used for peerreview papers.

\IEEEpeerreviewmaketitle

\section{Introduction}
\label{sec:introduction}
Currently, according to the report of International Data Corporation, there are nearly 3 billions smart-phones on the world \cite{DataCom,pandey2019incentivize}, which generate a huge amount of personal data. 
Nowadays, mobile devices equipped with specialized hardware architectures and computing engines can handle the machine learning problem effectively.
In addition, the application of machine learning techniques in mobile devices has grown rapidly. 
Furthermore, due to the limitation of wireless communication resources and privacy protection problem, the conventional central machine learning techniques, which upload all data of mobile devices to the central sever, are becoming less attractive. 
For this reason, federated learning (FL) is promoted, which is implemented distributively at the edge of the network \cite{bonawitz2019towards,le2020auction}.
In FL, mobile users can collaboratively train a global model using their own local data. 
Mobile users compute the updates of the current global model, and then send back the updates to the central server for aggregation and build a new global model. 
This process is repeated until an accuracy level of the global learning model is achieved. 
By this way, FL can preserve the personal information and data of mobile users. 
\textcolor{black}{In addition, FL will significantly promote services that have unparalleled versatile data collection and model training on a large scale.} Take the app Ware as an example. This application can help the users in avoiding  heavy traffic roads, but users have to share their own locations to the server. If FL is applied to this app, users only need to send the intermediate gradient values to the server rather than the raw data \cite{tran2019federated}.
Last but not least, the development of mobile edge computing provides an immense exposure to
extract the benefits of FL \cite{khan2020self,chen2020federated}.

In spite of the above mentioned benefits of FL, there are remaining challenges of having an efficient FL framework. Firstly, data samples per mobile device are small to train a high-quality learning model so a large number of mobile users are needed to ensure cooperation. 
In addition, the mobile users who join the learning process are independent and uncontrollable. 
Here, mobile users may not be willing to participate in the learning due to the energy cost incurred by model training. 
In other words, the base station (BS), which generates the global model, has to stimulate the mobile users for participation. 
Moreover, because the wireless resource is limited, the BS needs to allocate the resources reasonably to avoid the congestion, guarantee the model training performance and optimize the total utilities of the BS and mobile users.

To deal with the above challenges, in this paper, we model the FL service between the BS and mobile users as an auction game in which the BS is buyer and mobile users are sellers. 
In particular, the BS first initiates and announces a FL task. 
When each mobile user receives the FL task information,  they decide the amount of resources required to participate in the model training. 
After that, each mobile user submits a bid, which includes the required amount of resource, local accuracy, and the corresponding energy cost, to the BS. 
%The true cost is the minimal energy cost that the mobile user experiences corresponding with resource amount. 
%Based on submitted bids of mobile users, the BS selects different sets of participant users and service payment. 
%Finally, the BS coordinates the selected mobile users to conduct model training. 
%The winner selection in the proposed mechanism is based on primal dual greedy algorithm. 
\textcolor{black}{
	Moreover, the BS plays the role of auctioneer to decide the winners among mobile users as well as clear payment for the winning mobile users.  
	In addition, the auction used in this paper is a type of combinational auction \cite{nisan2007algorithmic,han2012game} since each mobile user can bid for combinations of resources. 
	However, the proposed auction mechanism allows mobile users sharing the resources at the BS, which is different from the conventional combinatorial auction.
	The proposed mechanism directly determines the trading rules between the buyer (BS) and sellers (mobile users) and motivates the mobile users to participate in the model training. 
	Compared with other incentive mechanism approaches (e.g.,  contract theory \cite{kang2019incentive}) in which the service market is a monopoly market, where mobile users can only decide whether or not to accept the contracts, the proposed auction enables mobile users to bids any combinations of resources.
	Moreover, the proposed auction mechanism can simultaneously provide truthfulness and individual rationality. 
	An auction mechanism is truthful if a bidder's utility does not increase when that bidder makes other bidding strategies, rather than the true value. 
	Revealing the true value is a dominant strategy for each participating user regardless of what strategies other users use\cite{wang2017enabling}. 
	An absent-truthfulness auction mechanism could leave the door to possible market manipulation and produce inferior results \cite{klemperer2002really}. 
	Additionally, if the value of any bidder is non-negative, an auction process will ensure individual rationality.}

The contributions of this paper are summarized as follows:
\begin{itemize}
	%\item First, we present a federated learning model in which cellular-connected wireless mobile users transmit their locally trained federated learning  models to a base station (BS) that generates the global federated learning model and transmits it back to the users. 
	%For the considered federated learning model, the bandwidth for uplink transmission is limited and, hence, the BS needs to select appropriate users to execute the federated learning algorithm so as to minimize the social cost.
	\item We propose an auction framework for the wireless FL services market. Then, we present the bidding cost in every user's bid submitted to the BS. From the perspective of mobile users, each mobile user make optimal decisions on the amount of resources and local accuracy so that the energy cost is minimized while delay requirement of FL is satisfied.
	\item From the perspective of the BS, we formulate the winner selection problem in the auction game as the social welfare maximization problem which is a NP-hard problem.
	We propose a primal-dual greedy algorithm to deal with the NP-hard problem in selecting the winning users and critical value based payment.
	We also proposed auction mechanism is truthful, individual rational and computational efficient.
	\item Finally, we carry out the numerical study to show that a proposed auction mechanism can guarantee the approximation factor of the integrality to the maximal welfare that is derived by the optimal solution and outperforms compared with baseline.
\end{itemize}
The rest of this paper is organized as follows.
Section~\ref{sec_relatedwork} summarizes the related work. 
The system model is introduced in Section~\ref{sec_system_model}.
We describe the problem formulation in Section~\ref{sec_bid}.
% The server provisioning is presented in Section~\ref{sec_auction}, followed by 
We present the auction-based resource purchasing mechanism in Section~\ref{sec_auction}.
Simulation results are given in Section~\ref{sec_simulation}.
Finally, Section~\ref{sec_conclusion} concludes the paper.

\section{Related Works}\label{sec_relatedwork}
Due to the resource constraints and the heterogeneity of mobile users, some focus issues are resource allocation, client selection and incentive mechanism to improve the efficiency of FL.
%Recently, several existing works studied the implementation problem of federated learning over wireless link \cite{bonawitz2019towards,chen2019joint,Pandey2019crowd}. 
The authors in \cite{tran2019federated} posed the joint learning and transmission energy minimization problem for FL. 
In this paper, all users upload their learning model to the BS in a synchronous manner.
The work in \cite{Yang2019} also considered the latency and energy consumption minimization problem for the case of asynchronous transmission.
The work in \cite{chen2019joint} explored the problem of reducing the learning loss function by considering packet errors over wireless links, but this research ignored the computation delay of the local learning model.
%The authors in  jointly consider the learning and communication problem. They aim to minimize a weighted sum of the completion time of learning, local computation energy, and transmission energy of all users, that captures the tradeoff of latency and energy consumption for learning. FDMA can be implemented in an asynchronous manner.
In \cite{Zeng2019}, the authors suggested energy-efficient strategies for allocating bandwidth and scheduling while at the same time, guaranteeing learning efficiency. 
The derived optimal policies allocate more bandwidth to those scheduled devices with weaker channels or lower computing capacities, which are the bottlenecks of synchronized model updates in FL. 
%In \cite{Samarakoon2019} the authors proposed a novel distributed approach based on federated learning to learn the network-wide queue dynamics in vehicular networks for achieving ultra-reliable low-latency communication (URLLC) via a joint power and resource allocation problem. 
%The participation of vehicles provides queue lengths related information to parameterize the distribution of extremes.

However, the works in \cite{tran2019federated,Yang2019,chen2019joint,Zeng2019} overlooked the problem of client selection to build a high-quality machine learning model. 
The authors in \cite{Nishio} designed a protocol called FedCS. 
The FedCS protocol has a resource request phase to gather information such as computing power and wireless channel states from a subset of randomly selected clients, i.e., FL workers, that are able to finish the local training punctually. 
%Compared with the protocol that ignores the client selection, the FedCS can achieve higher performance. 
%Besides improving the training efficiency, the authors in \cite{Li2019a} discussed the fairness issue that if a protocol selects the clients by the computing power, the final trained model would more cater to the data distribution of clients with high computational capability. 
A q-FedAvg training algorithm for selecting the client by the computational power was proposed in \cite{Li2019a}. 
This proposed algorithm in \cite{Li2019a} can improve the training efficiency and solve the fairness issue.
The study in \cite{wang2019edge} recommended combining deep reinforcement learning (DRL) and FL frameworks with mobile edge systems to optimize computing, caching and communication.
The study in \cite{yang2020federated} jointly considered the device selection and beamforming fast global model aggregation. 
They used the principle of over-the-air computation to exploit signal superposition multiple access channels.

Concerning the incentive mechanism design, the authors in \cite{Feng2019} proposed a Stackelberg game model to investigate the interactions between the server and the mobile devices in a cooperative relay communication network. 
The mobile devices determine the price per unit of data for individual profit maximization, while the server chooses the size of training data to optimize its own profit. 
The authors in \cite{kang2019incentive} studied the block-chained FL architecture and proposed the contract theory based payment mechanism to incentivize the mobile devices to take part in the FL. 
However, \cite{kang2019incentive} largely provided a latency analysis for the related applications.
The work in \cite{Pandey2019crowd} designed and analyzed a novel crowdsourcing framework to enable FL. 
In \cite{Pandey2019crowd}, a two-stage Stackelberg game model was adopted to jointly study the utility maximization of the participating clients and multi-access edge computing (MEC) server interacting via an application interface to construct a high-quality learning model. 
In \cite{zhan2020learning}, a Stackelberg game for FL in IoT was proposed to tackle the challenge of incentivizing people to join the FL by contributing their computational power and data. 
For the cases where the knowledge of participants' decisions and accurate contribution evaluation are accessible, the Nash Equilibrium was derived, and an algorithm based on DRL was built to unknown the knowledge of participants' decisions and accurate contribution evaluation for the cases.
However, both \cite{Pandey2019crowd} and \cite{zhan2020learning} studied only the uniform pricing scheme for participants.

Different from the Stackelberg game and contract theory, the auction mechanism allows mobile users to actively report its cost. Therefore, the BS is capable of understanding their status and requests adequately.
\cite{zeng2020fmore} adopted the multi-dimensional procurement auction to motivate nodes to participate in FL.
However, there exist some differences between \cite{zeng2020fmore} and our work: (a) \cite{zeng2020fmore} the bids submitted by edge node clarifies the combination of resources and the expected payment, which is based on the private cost parameters while in our work, the bid declares the combination of resource, local accuracy and the cost, which is determined based on latency and energy cost models; (b) the winner selection in \cite{zeng2020fmore} based on the scoring function announced by the aggregator while in our work, winners are selected in order to optimize social welfare and ensure resource efficiency. 

%In our work, we proposed the auction based incentive mechanism to mobile users in the federated learning service market.
\begin{figure*}
	\centering
	\includegraphics[width=1.05\linewidth]{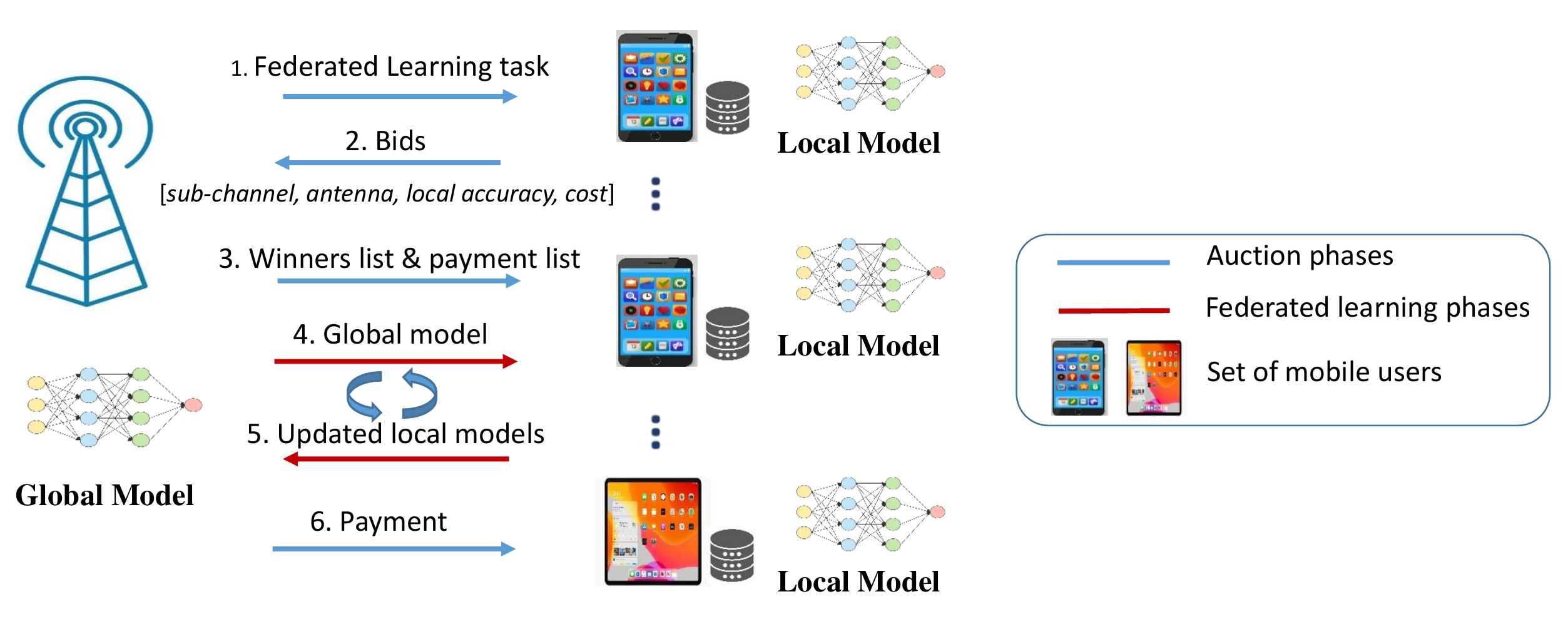}
	\caption{System model.}
	\label{fig_system_model}
	\vspace{0.5em}
\end{figure*}
\section{System Model: Federated Learning Services Market}\label{sec_system_model}
\subsection{Preliminary of Federated Learning}\label{subsec_fl}
Consider a cellular network in which one BS and a set $\mathcal{N}$ of $N$ users cooperatively perform a FL algorithm for model learning, as shown in Fig.~\ref{fig_system_model}.
Each user $n$ has $\boldsymbol{s}_n$ local data samples. Each data set $\boldsymbol{s}_n=\{a_{nk},b_{nk, 1 \leq k \leq s_n}\}$ where $a_{nk}$ is an input and $b_{nk}$ is its corresponding output.
%In this section, we present the preliminary of federated learning implemented over wireless network.
The FL model trained by the dataset of each user is called the local FL model, while the FL model at the BS aggregates the local model from all users as the global FL model. 
We define a vector $\omega$ as the model parameter. We also introduce the loss function $l_n(\omega, a_{nk},b_{nk})$ that captures the FL performance over input vector $a_{nk}$ and output $b_{nk}$. 
The loss function may be different, depending on the different learning tasks. The total loss function of user $n$ will be
\begin{equation}
L_n(\omega)=\frac{1}{s_n} \sum_{k=1}^{s_n} l_n(\omega,a_{nk},b_{nk}).
\end{equation}
Then, the learning model is the minimizer of the following global loss function minimization problem 
\begin{equation}\label{FLprob}
\min_{\omega} \quad L(\omega) = \sum_{n=1}^N \frac{s_n}{S}L_n(\omega)= \frac{1}{S} \sum_{n=1}^N \sum_{k=1}^{s_n} l_n(\omega,a_{nk},b_{nk}),
\end{equation}
where $S=\sum_{n=1}^N s_n$ is the total data samples of all users.

To solve the problem in (\ref{FLprob}), we adopt the FL algorithm of \cite{konevcny2016federated}.
The algorithm uses an iterative approach that requires a number of global iterations (i.e., communication rounds) to achieve a global accuracy level. 
In each global iteration, there are interactions between the users and BS. 
Specifically, at a given global iteration $t$, users receive the global parameter $\omega^{t}$, users computes $\triangledown L_n(\omega^{t}), \forall n$ and send it to the BS. The BS computes \textcolor{black}{\cite{Yang2019}}
\begin{equation}
\triangledown L(\omega^{t}) = \frac{1}{N} \sum_{n=1}^N \triangledown L_n(\omega^{t}),
\end{equation}
and then broadcasts the value of $\triangledown L(\omega^{t})$ to all participating users. 
Each participating user $n$ will use local training data $s_n$ to solve the local FL problem is defined as
\begin{equation}\label{FLprob2}
\begin{aligned}
&\min_{\phi_n} \quad \mathcal{G}_n(\omega^t, \phi_n)\\
& = L_n(\omega^t + \phi_n) - \left( \triangledown L_n(\omega^t) - \varpi \triangledown L(\omega^t)\right)^T  \phi_n, 
\end{aligned}
\end{equation} 
where $\phi_n$ represents the difference between global FL parameter and local FL parameter for user $n$.
Each participating user $n$ uses the gradient method to solve (\ref{FLprob2}) with local accuracy $\varepsilon_n$ that characterizes the quality of the local solution, and produces the output $\phi_n$ that satisfies
\begin{equation}
\mathcal{G}_n(\omega^t, \phi_{n}) -  \mathcal{G}_n(\omega^t, \phi^*_{n}) < \varepsilon_n  (\mathcal{G}_n(\omega^t, 0) -  \mathcal{G}_n(\omega^t, \phi^*_{n})).
\end{equation}
Solving (\ref{FLprob2}) also takes multiple local iterations to achieve a particular local accuracy.
Then each user $n$ sends the local parameter $\phi_n$ to the BS.
Next, the BS aggregates the local parameters from the users and computes 
\begin{equation}
\omega^{t+1}=\omega^{t}+\frac{1}{N}\sum_{n=1}^N \phi^t_{n},
\end{equation} 
and broadcasts the value to all users, which is used for next iteration $t+1$. This process is repeated until the global accuracy $\gamma$ of (\ref{FLprob}) is obtained. 

\textcolor{black}{With the assumption on $L_n(\omega)$}, the general lower bound on the number of global iterations is depends on local accuracy $\varepsilon$ and the global accuracy $\gamma$ as \cite{Yang2019}:
\begin{equation}\label{glo_iteration}
I^g(\gamma,\varepsilon)=\frac{C_1\log(1/\gamma)}{1-\varepsilon},
\end{equation}
where the local accuracy measures the quality of the local solution as described in the preceding paragraphs. 
%Further, in this formulation, we have replaced the term $O(log( 1/\gamma))$ in the numerator with $\varpi · log( 1/\gamma)$, for a constant $ \varpi > 0$. 

In (\ref{glo_iteration}), we observe that a very high local accuracy (small $\varepsilon$) can significantly boost the global accuracy $\gamma$ for a fixed number of global iterations $I^g$ at the BS to solve the global problem. 
However, each user $n$ has to spend excessive resources in terms of local iterations, $I^l_n$
to attain a small value of $\varepsilon_n$. The lower bound on the number of local iterations needed to
achieve local accuracy $\varepsilon_n$ is derived as \cite{Yang2019}
\begin{equation}
I^l_n(\varepsilon_n)=\vartheta_n \log \left( \frac{1}{\varepsilon_n} \right),
\end{equation}
where $\vartheta_n > 0$ is a parameter choice of user $n$ that depends on parameters of $L_n(\omega)$ \cite{Yang2019}. 
In this paper, we normalize $\vartheta_n=1$.
Therefore, to address this trade-off, the BS can setup an economic interaction environment to motivate the participating users to enhance local accuracy $\varepsilon_n$. 
Correspondingly, with the increased payment, the participating users are motivated to attain better local accuracy $\varepsilon_n$ (i.e., smaller values), which as noted in (\ref{glo_iteration}) can improve the global accuracy $\gamma$ for a fixed number of iterations $I^g$ of the BS to solve the global problem.
In this case, the corresponding performance bound in (\ref{glo_iteration}) for the heterogeneous responses $\epsilon_n$ can be updated to capture the statistical and system-level heterogeneity considering the worst response of the participating users as: 
%In this scenario, to capture the statistical and system-level heterogeneity, the corresponding performance bound in (\ref{glo_iteration}) for heterogeneous responses $\varepsilon_n$ can be modified considering the worst-case response of the participating users as
\begin{equation}
I^g(\gamma,\varepsilon_n)=\frac{\varpi\log(1/\gamma)}{1-\max_n\varepsilon_n}, \forall n.
\end{equation}

\subsection{Computation and Communication Models for Federated Learning}
%We denote $s_n$ as the local data samples that mobile user $n \in \mathcal{N}$ uses to participate in the federated learning task. 
%Each mobile user $n \in \mathcal{N}$ with a local training data set uses a size $s_n$ of its local data samples to participate in the federated learning task. 

The contributed computation resource that user $n$ contributes for local model training is denoted as $f_n$. 
Then, $c_n$ denotes the number of CPU cycles needed for the user $n$ to perform one sample of data in local training. 
Thus, energy consumption of the user for one local iteration is presented as 
\begin{equation}
E^{com}_n(f_n)= \zeta c_n s_n f_n^2,
\end{equation}
where $\zeta $ is the effective capacitance parameter of computing chipset for user $n$.
%Therefore, the time for computing of local iteration in local model training is $\frac{c_n s_n}{f_n}$. 
%\subsection{Communication Model for Federated Learning}
The computing time of a local iteration at the user $n$ is denoted by
\begin{equation}
T^{comp}_n=\frac{c_ns_n}{f_n}.
\end{equation}
It is noted that the uplink from the users to the BS is used to transmit the parameters of the local FL model while the downlink is used for transmitting the parameters of the global FL model. In this paper, we just consider the uplink bandwidth allocation due to the relation of the uplink bandwidth and the cost that user experiences during learning a global model.
%We use OFDMA 
We consider the uplink transmission of an OFDMA-based cellular system. A set of $\mathcal{B} = \{1, 2, . . . , B\}$ subchannels each with bandwidth $W$. Moreover, the BS is equipped with $A$ antennas and each user equipment has a single antenna (i.e., multi-user MIMO). We assume $A$ to be large (e.g., several hundreds) to achieve massive MIMO effect which scales up traditional MIMO by orders of magnitude.
Massive MIMO uses spatial-division multiplexing. 
%The transmission rate of mobile user is denoted as
The achievable uplink data rate of mobile user $n$ is expressed as \cite{hao2019energy}
\begin{equation}
r_n= b_{n}W \log_2 \left(1+ \frac{(A_n-1)p_nh_n}{b_{n}WN_0}\right),
\end{equation}
%\begin{equation}
%r_n^k=W \log (1+ \frac{p_nh_{n,k}}{WN_0}),
%\end{equation}
where $p_n$ is the transmission power of user $n$, $h_n$ is the channel gain of peer to peer link between user and the BS, $N_0$ is the background noise, $A_n$ is the number of antennas the BS assigns to user $n$, and $b_n$ is the number of sub-channels that user $n$ uses to transmit the local model update to the BS. 

We denote $\sigma$ as the data size of a local model update and it is the same for all users.  Therefore, the transmission time of a local model update is  
\begin{equation}
T^{com}_n(p_n,A_n,b_n)=\frac{\sigma}{r_n}.
\end{equation}  
To transmit local model updates in a global iteration, the user $n$ uses the amount of energy given as 
\begin{equation}
E^{com}(p_n,f_n,A_n,b_n)=T^{com} p_n = \frac{\sigma p_n}{r_n}.
\end{equation}
%\begin{equation}
%T^{com}_{n,k}=\frac{\sigma}{r_n^k}.
%\end{equation}
Hence, the total time of one global iteration for user $n$ is denoted as 
\begin{equation}\begin{aligned}
&T^{tol}_{n}(p_n,f_n,A_n,b_n,\varepsilon_n)\\
& = \log \left(\frac{1}{\varepsilon_n}\right) T^{comp}_n(f_n) + T^{com}_{n}(p_n,A_n,b_n).
\end{aligned}
\end{equation}
%\begin{equation}
%E^{com}_{n,k}=T^{com}_{n,k} p_n = \frac{\sigma p_n}{r_n^k}.
%\end{equation}
Therefore, the total energy consumption of a user $n$ in one global iteration is denoted as follows 
\begin{equation}\begin{aligned}
&E^{tol}_{n}(p_n,f_n,A_n,b_n,\varepsilon_n)\\
&= \log \left(\frac{1}{\varepsilon_n}\right) E^{comp}_n(f_n) + E^{com}_{n}(p_n,A_n,b_n).
\end{aligned}
\end{equation}
\subsection{Auction Model}

As described in Fig. 1, the BS first initializes the global network model. 
Then, the BS announces the auction rule and advertises the FL task to the mobile users. 
The mobile users then report their bids. 
Here, mobile user $n$ submits a set of $I_n$ of bids to the BS. A bid $\Delta_{ni}$ denotes the $i$th bid submitted by the mobile user $n$. Bid $b_{ni}$ consists of the resource (sub-channel number $b_{ni}$, antenna number $A_{ni}$, local accuracy level $\varepsilon_{ni}$) and the claimed cost $v_{ni}$ for the model training. Each mobile user $n$ has its own discretion to determine its true cost $V_{ni}$, which will be presented in Section \ref{sec_bid}. Let $x_{ni}$ be a binary variable indicating the bid $\Delta_{ni}$ wins or not. 
After receiving all the bids from mobile users, the BS decides winners and then allocates the resource to the winning mobile users. The winning mobile users join the FL and receive the payment after finishing the training model.
\section{Deciding mobile users's bid}\label{sec_bid}
To transmit the local model update to the BS, mobile users need sub-channels and antennas resources.
However, given the maximum tolerable time of FL, there is a correlation between resource and corresponding energy cost. 
In this section, we present the way mobile users decide bids. Specially, for bid $\Delta_{ni}$, mobile user $n$ calculates transmission power $p_{ni}$, computation resource $f_{ni}$ and cost $v_{ni}$ corresponding to a given sub-channel number $b_{ni}$ and antenna number $A_{ni}$. However, the process to decide mobile users' bid is the same for every submitted bids. Thus, we remove the bid index $i$ in this section. 
The energy cost of mobile user $n$ is defined as 
\begin{subequations} \begin{align}   
	\bold{P1:}\quad \underset{f_n, p_n, A_n, b_n,\varepsilon_n}{\min}\quad 
	& I_0^n E^{tol}_n(p_n,f_n,A_n,b_n,\varepsilon_n)\\    
	s.t.\quad 
	& I_0^n T^{tol}_n(p_n,f_n,A_n,b_n,\varepsilon_n) \leq T_{max}, \label{consp1}\\ 
	& f_{n} \in [f^{\min}_n, f^{\max}_n],\\
	& p_{n} \in (0, p^{\max}_n],\\
	& \varepsilon_n \leq (0,1],\\
	& A_{n} \in (0, A^{\max}_n],\\
	& b_{n} \in (0, b^{\max}_n], 
	\end{align}   
\end{subequations}
%\begin{subequations} \begin{align}   
%	\textbf{P1:}\quad \underset{x}{\min}\quad 
%	&\sum_{k,n} x_{n,k}( E^t_{k,n} +  T^t_{k,n}) \\    
%	s.t.\quad 
%	& \sum_k x_{n,k} \leq 1, \forall n\\
%	&\sum_n x_{n,k} \leq 1, \forall k\\
%	& x_{n,k} \in \{0,1\}
%	\end{align}   %\end{subequations}
where $f^{\max}_n$ and $p^{\max}_n$ are the maximum local computation capacity and maximum transmit power of mobile user $n$, respectively. $A^{\max}_n$ and $b^{\max}_n$ are the maximum antenna and maximum sub-channel that mobile user $n$ can request in each bid, respectively. $A^{\max}_n$ and $b^{\max}_n$ are chosen by mobile user $n$. $I_0^n=\frac{C_1\log(1/\gamma)}{1-\varepsilon_n}$ is the lower bound of the number global iterations corresponding to local accuracy $\varepsilon_n$. Note that the cost to the mobile user cannot be the same over iterations. However, to make the problem more tractable, we consider minimizing the approximated cost rather than the actual cost, similar to approach in \cite{Pandey2019crowd, wang2018edge}. Constraint (\ref{consp1}) indicates delay requirement of FL task.

According to \textbf{P1}, the maximum number of antennas and sub-channels are always energy efficient, i.e., the optimal antenna is $A_n=A_n^{max}, b_n=b_n^{max}$ and $\varepsilon^*_n, p^*_n,f^*_n$ are the optimal solution to:
\begin{equation} \begin{aligned}   
\bold{P2:}\quad \underset{f_n,p_n,\varepsilon_n}{\min}\quad 
& I_0^n E^{tol}_n(p_n,f_n,\varepsilon_n)\\    
s.t.\quad 
& I_0^n T^{tol}_n(p_n,f_n,\varepsilon_n)  \leq T_{max},\\ 
&  f_{n} \in [f^{\min}_n, f^{\max}_n],\\
& \varepsilon_{n} \in (0, 1],\\
& p_{n} \in (0, p^{\max}_n].
\end{aligned}   
\end{equation}
Because of the non convexity of \textbf{P2}, it is challenging to obtain the global optimal solution. To overcome the challenge, an iterative algorithm with low complexity is proposed in the following subsection.
\subsection{Iterative Algorithm}
The proposed iterative algorithm basically involves two steps in each iteration. To obtain the optimal, we first solve (\textbf{P2}) with fixed $\varepsilon_n$, and then $\varepsilon_n$ is updated based on the obtained $f_n,p_n$ in the previous step. %The advantage of this iterative algorithm lies in that we can obtain solution of $f_n,p_n$ and $\varepsilon_n$ in each step.
In the first step, we consider the first case when $\varepsilon_n$ is fixed, and \textbf{P2} becomes
\begin{equation} \begin{aligned}   
\bold{P3:}\quad \underset{f_n,p_n,\varepsilon_n}{\min}\quad 
& I_0^n E^{tol}_n(p_n,f_n,\varepsilon_n)\\    
s.t.\quad 
& I_0^n T^{tol}_n(p_n,f_n,\varepsilon_n)  \leq T_{max},\\ 
&  f_{n} \in [f^{\min}_n, f^{\max}_n],\\
& p_{n} \in (0, p^{\max}_n].
\end{aligned} 
\end{equation}
%Note that in \textbf{P2}, the allocation of the uplink transmission power $p$ and computation resources $f$ are coupled from each other in the constraint (\ref{consp1}). However, \textbf{P2} can be solved by optimizing communication and computation resources in an iterative manner.
\textbf{P3} can be decomposed into two sub-problems as follows.
\subsubsection{Optimization of Uplink Transmission Power}
Each mobile user assigns its transmission power by solving the following problem:
\begin{equation} \begin{aligned}   
\bold{P3a:}\quad \underset{p_n}{\min}\quad 
&  f(p_n)\\    
s.t.\quad 
& I_0^n\left(f(p_n)/\sigma+T^{comp}_n\right) \leq T_{max},\\ 
&  p_{n} \in (0, p^{\max}_n],\\
& f_n, \varepsilon_n\,\,\ \text{are given}.
\end{aligned}   
\end{equation}
where $f(p_n)=\frac{\sigma p_n}{b_n W \log_2(1+\frac{(A_n-1)p_nh_n}{b_nWN_0})}$.
\setlength{\textfloatsep}{0pt}
\begin{algorithm}[t]
	\caption{Optimal Uplink Power Transmission} \label{alg1}
	Calculate $\phi(p_n^{max})$\\
	Calculate $p_n^{min}$ so that $T^t_n(p_n^{min}) = T_{max}$\\
	\eIf{$\phi(p_n^{max} < 0)$}
	{ $p_n^{*}=p_n^{max}$
	}
	{ $p_1=\max(0,p_n^{min})$ and $p_2=p_n^{max}$\\
		\While{($p_2 - p_1 \leq \epsilon$) }
		{ $p_u=(p_1 + p_1)/2$ \\
			\eIf{$\phi(p_u) \leq 0 $}{\nl $p_1=p_u$\\}{\nl $p_2=p_u$\\}
		}
		$p_n^{*}=(p_1 + p_2)/2$	\\
	}	
\end{algorithm}	
%\begin{lemma}
Note that $f(p_n)$ is quasiconvex in the domain \cite{lyu2016multiuser}.
%\end{lemma}
A general approach to the quasiconvex optimization problem is the bisection method, which solves a convex feasibility problem each time \cite{boyd2004convex}. 
However, solving convex feasibility problems by an interior cutting-plane method requires $O(\kappa^2/\alpha^2)$ iterations, where $\kappa$ is the dimension of the problem \cite{lyu2016multiuser}.
On the other hand, we have
\begin{equation}
f'(p_n)=\frac{\sigma \log_2(1+\theta_np_nh_n) + \frac{\sigma p_n \theta_n h_n}{\ln2 (1+\theta_np_nh_n)}}{{b_n W( \log(1+\theta_np_nh_n)})^2},
\end{equation}
where $\theta_n=\frac{(A_n-1)}{WN_0}$. Then, we have 
\begin{equation}
\phi(p_n)=\sigma \log_2(1+\theta_np_nh_n) + \frac{\sigma p_n \theta_n h_n}{\ln2 (1+\theta_np_nh_n)}
\end{equation}
is a monotonically increasing transcendental function and negative at the starting point $p_n = 0$ \cite{lyu2016multiuser}.
Therefore, in order to obtain the optimal power allocation $p_n$ as shown in Algorithm \ref{alg1}, we follow a low-complexity bisection method by calculating $\phi(p_n)$ rather than solving a convex feasibility problem each time.
\subsubsection{Optimization of CPU cycle frequency and number of antennas:}

\begin{equation} \begin{aligned}   
\bold{P3b:}\quad \underset{f_n}{\min}\quad 
&  I_0^n\log \left(\frac{1}{\varepsilon_n} \right)\zeta c_n s_n f_n^2\\    
s.t.\quad 
& I_0^n\left(\log \left(\frac{1}{\varepsilon_n} \right)\frac{c_ns_n}{f_n}+T^{com}_n\right)\leq T_{max},\\ 
& f_{n} \in [f^{\min}_n, f^{\max}_n],\\
& p_n, \varepsilon_n \,\,\ \text{are given}.
\end{aligned}   
\end{equation}
\textbf{P3b} is the convex problem, so we can solve it by any convex optimization tool.

In the second step, \textbf{P2} can be simplified by using $f_n$ and $p_n$ calculated in the first step as: 
\begin{subequations}\begin{align}
	\bold{P4:}\quad \underset{\varepsilon_n}{\min}\quad
	& \frac{\gamma_1 \log_2(1/\varepsilon_n)+\gamma_2}{1-\varepsilon_n}\\
	s.t.\quad 
	& T^{tol}_n \leq T_{max}, \label{P5_c}
	\end{align}
\end{subequations}
where
$\gamma_1=a E^{comp}_n$ and $\gamma_2 = a E^{com}_n$. 
The constraint (\ref{P5_c}) is equivalent to $T_n^{com} \leq \vartheta(\varepsilon_n)$, where $\vartheta(\varepsilon_n)=\frac{1-\varepsilon_n}{m}T_{max}+\frac{c_ns_n\log_2\varepsilon_n}{f_n}$. 
We have $\vartheta(\varepsilon_n) '' < 0 $, and therefore, $\vartheta(\varepsilon_n)$  is a concave function.
Thus, constraint (\ref{P5_c}) can be equivalent transformed to $\varepsilon_n^{min} \leq \varepsilon_n \leq \varepsilon_n^{max}$, where $\vartheta(\varepsilon_n^{min})=\vartheta(\varepsilon_n^{max})=T_n^{com}$.
Therefore, $\varepsilon_n$ is the optimal solution to
\begin{equation}
\begin{aligned}
\bold{P5:}\quad \underset{\varepsilon_n}{\min}\quad
& \frac{\gamma_1 \log_2(1/\varepsilon_n)+\gamma_2}{1-\varepsilon_n}\\
s.t.\quad 
& \varepsilon_n^{min} \leq \varepsilon_n \leq \varepsilon_n^{max}.
\end{aligned}
\end{equation}
\setlength{\textfloatsep}{0pt}
\begin{algorithm}[t]
	\caption{Optimal Local Accuracy} \label{alg3}
	Initialize $\varepsilon_n=\varepsilon_n^{(0)}$, set $j=0$\\
	\Repeat{$|H(\xi^{(n+1)})|/|H(\xi^{(n)})| < \epsilon_2$}
	{    Calculate $\varepsilon_n^{*}=\frac{\alpha_1}{(\ln2)\xi^j}$ \\
		Update $\xi^{(j+1)}=\frac{\gamma_1 \log_2(1/\varepsilon_n)+\gamma_2}{\varepsilon_n}$\\
		Set $j=j+1$\\}    
\end{algorithm}		

\setlength{\textfloatsep}{0pt}
\begin{algorithm}[t]
	\caption{Iterative Algorithm} \label{alg4}
	Initialize a feasible solution $p_n,f_n,\varepsilon_n$ and set $j=0$.\\
	\Repeat{Objective value of \textbf{P2} converges}
	{    With $\varepsilon_n^{(j)}$ obtain the optimal $p_n^{(j+1)},f_n^{(j+1)}$ of problem \\
		With $p_n^{(j+1)},f_n^{(j+1)}$ obtain the optimal $\varepsilon_n^{(j+1)}$ of problem   \\
		Set $j=j+1$\\}    
\end{algorithm}
Obviously, the objective function of \textbf{P5} has a fractional in nature, which is generally difficult to solve. 
According to \cite{dinkelbach1967nonlinear, Yang2019}, solving \textbf{P5} is equivalent to finding the root of the nonlinear function $H(\xi)$ defined as follows
\begin{equation}\label{Dinkelbach}
H(\xi)=\min_{\varepsilon_n^{min} \leq \varepsilon_n \leq \varepsilon_n^{max}}\gamma_1 \log_2(1/\varepsilon_n)+\gamma_2-\xi(1-\varepsilon_n) 
\end{equation}
Function $H(\xi)$ with fixed $\xi$ is convex. Therefore, the optimal solution $\varepsilon_n$ can be obtained by setting the first-order derivative of $H(\xi)$ to zero, which leads to the optimal solution is $\varepsilon^*_n=\frac{\gamma_1}{(\ln2\xi)}$. Thus, similar to \cite{Yang2019}, problem \textbf{P5} can be solved by using the Dinkelbach method in \cite{dinkelbach1967nonlinear} (shown as Algorithm \ref{alg3}).

The algorithm that solves problems \textbf{P2} is given in Algorithm \ref{alg4}, iteratively solving problems \textbf{P3} and \textbf{P4}. 
%By iteratively solving problem \textbf{P3} and problem \textbf{P4}, the algorithm that solves problem \textbf{P2} is given in Algorithm \ref{alg4}. 
Since the optimal solution of problem \textbf{P3} and \textbf{P4} is obtained in each step, the objective value of problem \textbf{P2} is non-increasing in each step. Moreover, the objective value of problem \textbf{P2} is lower bounded by zero. Thus, Algorithm \ref{alg4} always converges to a local optimal solution.
\subsection{Complexity Analysis}
To solve the general energy-efficient resource allocation problem \textbf{P2} using Algorithm 3, the major complexity in each step lies in solving problems \textbf{P3} and \textbf{P4}. 
To solve problem \textbf{P3}, the complexity is $O(L_{e}\log_2(1/\epsilon_1))$, where $\epsilon_1$ is the accuracy of solving \textbf{P3} with the bisection method and $L_{e}$ is the number of iterations for optimizing $f_n$ and $p_n$. 
To solve problem \textbf{P4}, the complexity is $O(\log_2(1/\epsilon_2))$ with accuracy $\epsilon_2$ by using the Dinkelbach method.
As a result, the total complexity of the proposed Algorithm \ref{alg4} is $H_{e}S$, where $H_{e}$ is the number of iterations for problems \textbf{P3} and \textbf{P4} and $S$ is equal to $O(L_{e}\log_2(1/\epsilon_1))+O(\log_2(1/\epsilon_2))$.  

After deciding the bids, the mobile users submit bids to the BS. The following section describes the auction mechanism between the BS and mobile users for selecting winners, allocating bandwidth and deciding on payment. 
\section{Auction mechanism between BS and mobile users}\label{sec_auction}

\subsection{Problem Formulation}
In bid $\Delta_{ni}$ that mobile user $n$ submits to the BS includes the number of subchannels $b_{ni}$, the number of antennas $A_{ni}$, local accuracy $\epsilon_{ni}$, and claimed cost $v_{ni}$. 
The utility of one bid is the difference between the payment $g_{ni}$ and the real cost $V_{ni}$.
\begin{equation}
U_{ni}=
\begin{cases}
g_{ni}-V_{ni}, & \text{if bid $\Delta_{ni}$ wins}, \\
0, & \text{otherwise.}
\end{cases}
\end{equation}

The payment that the BS pays for winning bids is $\sum_{n, i} g_{ni}$.
%When users helps the BS to train the learning model, the BS can save the money. When the local accuracy increases, the number of global iterations decreases. Therefore, we model the BS' money saving due to the helping of user $n$ is a function of claimed local accuracy $\varepsilon_n$.
As we described in Section \ref{subsec_fl}, high local accuracy will significantly improve the global accuracy for a fixed number of global iterations.
%As we mentioned in Section \ref{subsec_fl}, for fixed number global iterations a very high local accuracy can remarkably improve the global accuracy.
The utility of the BS is the difference between the BS's satisfaction level and the payment for mobile users. The satisfaction level of the BS to bid $\Delta_{ni}$ is measured based on the local accuracy that mobile user $n$ can provide in the $i$th bid and is defined as follows 
%Therefore, the utility of the BS due to the training participation of user $n$ is defined as 
\begin{equation}
\chi_{ni}=\tau \varepsilon_{ni}.
\end{equation}
Thus, the total utilities of the system or the social welfare is 
\begin{equation}
\sum_{n, i}(\chi_{ni}- v_{ni})x_{ni}.
\end{equation}
If mobile users truthfully submit their cost, $V_{ni}=v_{ni}$, we have the social welfare maximization problem defined as follows:
\begin{subequations} \begin{align}   \label{prob_socialcost}
	\bold{P6:}\quad \underset{x}{\max}\quad 
	& \sum_{n, i}  (\chi_{ni}- v_{ni})x_{ni}\\    
	s.t.\quad 
	& \sum_{n} x_{ni}b_{ni} \leq B_{max}, \label{cons1}\\
	&  \sum_{n} x_{ni}A_{ni} \leq A_{max}, \label{cons2}\\
	&  \sum_{i} x_{ni} \leq 1,\forall n, \label{cons3}\\
	&  x_{ni} = \{0,1\} \label{cons4},
	\end{align}   
\end{subequations}
%where $c_k$ denotes for one subchannel of set $\mathcal{C}$. 
%The objective function of \textbf{P4} is to maximize the social welfare which is the sum of the real cost of bids selected by the BS. 
where (\ref{cons1}) and (\ref{cons2}) indicate the bandwidth resource (i.e., sub-channels)  and  the antennas limitation constraints of the BS, respectively. 
%The first constraint (\ref{cons1}) indicates the sub-channel number limitation of BS, the second constraint (\ref{cons2}) indicates the antenna number limitation of BS. 
Then, (\ref{cons3}) shows that a mobile user can win at most one bid and (\ref{cons4}) is the binary constraint that presents whether bid $\Delta_{ni}$ wins or not.

Problem \textbf{P6} is a minimization knapsack problem, which is known to be NP-hard. This implies that no algorithm is able to find out the optimal solution of \textbf{P6} in polynomial time. It is also known that a mechanism with Vickrey-Clarke-Groves (VCG) payment rule is truthful only when the resource allocation is optimal. Hence, using VCG payment directly is unsuitable due to the problem \textbf{P6} is computationally intractable.
To deal with the NP-hard problem, we proposed the primal-dual based greedy algorithm. The following economic properties are desired.

\textbf{Truthfulness:}  An auction mechanism is truthful if and only if for every bidder $n$ can get the highest utility when it reports true value.

\textbf{Individual Rational:} If each mobile user reports its true information (i.e., cost and local accuracy), the utility for each bid is nonnegative, i.e., $U_{ni} \geq 0$.

\textbf{Computation Efficiency:} The problem can be solved in polynomial time.

Among these three properties, truthfulness is the most challenging one to achieve. In order to design a
truthful auction mechanism, we introduce the following definitions.

\begin{defn}
	(Monotonicity): If mobile user $n$ wins with the bid $\Delta_{ni}=\{v_{ni},\varepsilon_{ni},b_{ni},A_{ni}\} $, then mobile user $n$ can win the bid with $\Delta_{nj}=\{v_{nj},\varepsilon_{nj},b_{nj},A_{nj}\} \succ \Delta_{ni}=\{v_{ni},\varepsilon_{ni},b_{ni},A_{ni}\} $.
\end{defn} 

The notation $\succ$ denotes the preference over bid pairs.
Specifically, $\Delta_{nj}=\{v_{nj},\varepsilon_{nj},b_{nj},A_{nj}\} \succ \Delta_{ni}=\{v_{ni},\varepsilon_{ni},b_{ni},A_{ni}\} $ if $\varepsilon_{nj}>\varepsilon_{ni}$ for $v_{nj}=v_{ni},b_{nj}=b_{ni},A_{nj}=A_{ni} $ or $v_{nj}<v_{ni},b_{nj}<b_{ni},A_{nj}<A_{ni} $ for $\varepsilon_{nj}=\varepsilon_{ni}$. 
The monotonicity implies that the chance to obtain a required bundle of resources can only be enhanced by either increasing the local accuracy or decreasing the amount of resources required or decreasing the cost.
\textcolor{black}{
	\begin{defn}
		(Critical Value): For a given monotone allocation scheme, there
		exists a critical value $c_{ni}$ of each bid $\Delta_{ni}$ such
		that $\forall n,i (\chi_{ni}-v_{ni}) \geq c_{ni}$ will be a winning bid, while $\forall n,i (\chi_{ni}-v_{ni}) < c_{ni}$ is a losing bid.
	\end{defn}
	In our proposed mechanism, the difference between the satisfaction based on local accuracy and cost of one bid can be considered as the value of that bid. 
	Therefore, the critical value can be seen as the minimum value that one bidder has to bid to obtain the requested bundle of resources. With the concepts of monotonicity and critical value, we have the following lemma. %\cite{mu2008truthful}
	\begin{lemma}\label{critical}
		An auction mechanism is truthful if the allocation scheme is monotone and each winning mobile user
		is paid the amount that equals to the difference between the satisfaction based on the local accuracy and the critical value.
	\end{lemma}
	\begin{IEEEproof}
		Similar Lemma 1 and Theorem 1 in \cite{wang2017enabling}. 
	\end{IEEEproof}}
	In the next subsection, we propose a primal-dual greedy approximation algorithm for solving problem \textbf{P6}. The
	algorithm iteratively updates both primal and dual variables and the approximation analysis is based on duality property. As the result, we firstly relax $1 \geq x_{ni} \geq 0$ of \textbf{P6} to have the linear programming relaxation (LPR) of \textbf{P6}. Then, we introduce the dual variable vectors $\boldsymbol{y}$, $\boldsymbol{z}$ and $\boldsymbol{t}$  corresponding to constraints (\ref{cons1}), (\ref{cons2}) and (\ref{cons3}) and we have the dual of problem LPR of \textbf{P6} can be written as
	\begin{subequations} \begin{align}   \label{prob_social cost_dual}
		\bold{P7:}\quad \underset{\bold{y,z,t}}{\max}\quad 
		&  \sum_{n \in \mathcal{N}} y_n +  zB_{max} + tA_{max}\\    
		s.t.\quad 
		& y_n + zA_{ni} + tB_{ni} \geq q_{ni}, \forall n,i , \label{cons1_dual}\\
		&  y_{n} \geq 0, \forall n, \label{cons2_dual}\\
		&  z,t \geq 0. \label{cons3_dual}
		\end{align}   
	\end{subequations} 
	In Section \ref{subsec_Algo_design}, we devise an greedy approximation algorithm and Section \ref{subsec_Algo_analysis}, a theoretical bound is achieved for the approximation ratio of the proposed algorithm.
	\subsection{Approximation Algorithm Design}\label {subsec_Algo_design}
	In this section, we use a greedy algorithm to solve problem \textbf{P6} %and at the same time, the method of dual fitting is used to derive an upper bound $\alpha$ via a linear programming (LP)-dual theory analysis. 
	The main idea of the greedy algorithm is to allocate the resource to bidders with the larger normalized value. 
	Specifically, after collecting all the bids from the mobile users, the BS as the auctioneer sorts the bids in a decreasing order of $\frac{q_{ni}}{s_{ni}}$ which is viewed as the normalized value of a bid, where $s_n^{i} = \eta_b B_n + \eta_a A_n$ is a weighted sum of the number of different types of resources requested and $q_{ni}=\chi_{ni}- v_{ni}$ is value of bid $\Delta_{ni}$. 
	\begin{algorithm}[t]
		\caption{The Greedy Approximation Algorithm} \label{alg}
		\textbf{Input:} $(B,A,\chi,v,B_{max},A_{max})$\\
		\textbf{Output:} solution $\textbf{x}$\\ 	
		$\mathcal{U} = \varnothing$, $\textbf{x}=\textbf{0} $\\
		$\forall n: y_n=0, \psi=0$;\\ 
		$\varphi=0, B=0, A=0 $;\\
		$s^i_n=\eta_b B_{ni} + \eta_a A_{ni}$;\\
		$q_{kj}=\chi_{ni}-v_{ni} $;\\
		\For{$n \in \mathcal{N} $}{
			$i_n=\arg \max_i \{q_{ni}\}$; \\		
		}
		
		$\kappa= \max \frac{s_{ni}}{s_{ni'}}$;\\
		%	\nl Re-index all bid pairs such that \\
		%	 $\frac{\chi^i_1-v^i_1}{s^i_1} \geq \frac{\chi^i_2-v^i_2}{s^i_2} \geq ...\geq \frac{\chi^i_K-v^i_K}{s^i_K} $\\
		\While{$\mathcal{N} \neq \varnothing$}
		{
			$\mu = \arg \max_{n \in \mathcal{N}} \frac{q_{ni}}{s_{n i_n}}$;\\
			\eIf{$B+b_{\mu i_\mu}<=B_{max}$ \text{and} $A+a_{\mu i_\mu}<=A_{max}$}
			{ $x_{\mu i_\mu}=1;\quad y_\mu=q_{\mu i_\mu}$;\\ 
				$\varphi =\varphi + q_{\mu i_\mu}; $\\
				$\psi=\frac{\sum_{n \in \mathcal{U}} q_{n i_n}}{\sum_{n \in \mathcal{U}}s_{n i_n}}$;\\ 
				$\mathcal{U} =\mathcal{U} \cup \{\mu\}$ \text{and} $\mathcal{N} =\mathcal{N} \setminus \{\mu\}$
			}
			{ \text{break;}}				
		}
		$\bar{\psi}=\kappa \psi$;\\
		$z = \eta_b \bar{\psi}$, $\quad t = \eta_a \bar{\psi}$
	\end{algorithm}	
	\subsection{Approximation Ratio Analysis}\label {subsec_Algo_analysis}
	In this subsection, we analyze approximation ratio of Algorithm \ref{alg}. 
	Our approach is to use the duality property to derive a bound for approximation algorithm.
	We denote the optimal solution and the optimal value of LPR of \textbf{P6} as $x^{*}_{ni}$ and $OP_f$.
	Furthermore, let $OP$ and $\varphi$ as the optimal value of \textbf{P6} and the primal value of \textbf{P6} obtained by Algorithm \ref{alg}.
	Our analysis consists of two steps. First, Theorem \ref{t2} shows that Algorithm \ref{alg} generates a feasible solution to \textbf{P7}, and Proposition \ref{prop1} provides approximation factor.
	\begin{theorem}\label{t2}
		Algorithm ~\ref{alg} provides a feasible solution to \textbf{P7}.
	\end{theorem}
	\begin{IEEEproof}
		We discuss the following three cases:
		\begin{itemize}
			\item Case 1: mobile user $\mu$ wins, i.e., 
			$\mu \in \mathcal{U}$ and $ b_{\mu i_{\mu}} = \text{max}_{i' \in \mathcal{I}_\mu} \{q_{\mu i'}\}$.
			Then we have $y_\mu=q_{\mu i_\mu} \geq q_{\mu i'}, \forall i' \in \mathcal{I}_\mu$.
			Thus, constraint \eqref{cons1_dual} is satisfied for all mobile users in $ \mathcal{U}$.
			
			\item Case 2: mobile user $\mu$ loses the auction, i.e.,  $\mu \in \mathcal{N} \setminus \mathcal{U}$. According to the while loop, it is evident that
			\begin{equation*}
			\frac{q_{n i_n}}{s_{ni_n}} > \frac{q_{\mu i_\mu}}{s_{\mu i_\mu}}, \forall n \in \mathcal{U}.
			\end{equation*}
			Therefore, $\psi > \frac{q_{\mu i_\mu}}{s_{\mu i_\mu}}$. Thus,
			\begin{equation*}
			\bar{\psi} \geq \kappa \frac{q_{\mu i_\mu}}{s_{\mu i_\mu}} \geq \frac{q_{\mu i_\mu}}{s_{\mu i_\mu}}.
			\end{equation*}
			In addition, we have
			\begin{equation*}
			q_{\mu i_\mu} \geq q_{\mu i'}\quad  \text{and} \quad \kappa>\frac{s_{\mu i_\mu}}{s_{\mu i'}}, \forall i' \neq i_\mu.
			\end{equation*} 
			Therefore, 
			\begin{equation*}
			\bar{\psi} \geq \frac{q_{\mu i'}}{n_{\mu i'}}, \forall i' \neq i_\mu.
			\end{equation*}
			Therefore, we have
			\begin{equation*}
			\eta_b\bar{\psi}B_{in} + \eta_a\bar{\psi}A_{in} \geq q_{in}, \forall i' \neq i_\mu.
			\end{equation*} 
			or 
			\begin{equation*}
			zC_{in} + tA_{in} \geq q_{in}, \forall i' \neq i_\mu.
			\end{equation*} 		
			Therefore, constraint \eqref{cons1_dual} is also satisfied for all mobile users in $ \mathcal{N} \setminus \mathcal{U}$.
		\end{itemize}
		\vspace{-0.1in}	
	\end{IEEEproof}	
	\begin{prop}\label{prop1}
		The upper bound of integrality gap $\alpha$ between \textbf{P6} and its relaxation and the approximation ratio of Algorithm \ref{alg} are $1+\frac{\kappa\Upsilon}{\Upsilon-S}$, where $\Upsilon=\eta_b B_{max}+ \eta_a A_{max}, S = \max_{n,i} s_{ni}.$
	\end{prop}
	
	\begin{IEEEproof}
		%In Lemma 1, we prove that $\phi$ and $\bar{\psi}$ are feasible solutions for the dual formulation. Therefore, according to LP duality, $\sum_{k=1}^{K} \phi_k + W\, \psi $ is the upper bound for the primal formulation \textbf{LPR}. 
		Let $OP$ and $OP_{f}$ be the optimal solution for \textbf{P6} and LPR of \textbf{P6}. %(\ref{prob2}), (\ref{probL}). 
		We can obtain the following:
		\begin{equation*}
		\begin{aligned}
		OP \leq OP_f &\leq \sum_{n=1}^{N} y_n +  zB_{max} + tA_{max}\\
		&\leq \sum_{n=1}^{N} y_n +  \bar{\psi}(\eta_b B_{max}+ \eta_a A_{max}) \\
		&\leq \sum_{n \in \mathcal {N}} q_{n i_n} + \bar{\psi}(\eta_b B_{max}+ \eta_a A_{max})\\
		%&\leq \sum_{k \in \mathcal {U}} b_{k j_k} + (\omega C_{max}+(1-\omega)A_{max}) \epsilon \frac{\sum_{k \in \mathcal{U}} b_{kj_k}}{\sum_{k \in \mathcal{U}}n_{kj_k}}\\
		&\leq \left(\sum_{n\in \mathcal {N}} q_{n i_n}\right)\left(1+\frac{(\eta_b B_{max}+ \eta_a A_{max}) \kappa}{\eta_b B_{max}+\eta_a A_{max}-S}\right)\\
		& \leq \varphi \left(1+\frac{\Upsilon \kappa}{\Upsilon-S}\right),
		\end{aligned} 
		\end{equation*}
		%$\text{} \leq \text{}$
		%where $\varphi$ is the social welfare of the near-optimal integer solution computed by the proposed greedy approximation algorithm. 
		\vspace{-0.1in}
		Therefore, the integrality $\alpha$ is given as
		\begin{equation*}
		\begin{aligned}
		& OP_f/OP\\
		\leq \quad & OP_f/\varphi\\
		\leq \quad & \left(1+\frac{\kappa\Upsilon}{\Upsilon-S}\right).
		\end{aligned} 
		\end{equation*}
		\vspace{-0.1in}
		The approximation ratio is	
		\begin{equation*}
		OP/\varphi \leq OP_f/\varphi \leq \left(1+\frac{\kappa\Upsilon}{\Upsilon-S}\right).
		\end{equation*} 
		\vspace{-0.1in}
	\end{IEEEproof}	
	\vspace{-0.1in}
	\subsection{Payment}
	
	Then we will find the critical value which is the minimum value a bidder has to bid to win the requested bundle of resources. In this paper, we consider the bid combinations submitted by mobile user $n$ as the combinations of bid submit by virtual bidders, in which each virtual bidder can submit one bid. Therefore, the number of virtual bidders corresponding to mobile user $n$ is equal to the number of bids $I_n$ that mobile user $n$ submits.
	Denote by $m$ the losing mobile user with the highest normalized value if mobile user $n$ is not participating in the
	auction. 
	Accordingly, the minimum value mobile user $n$ needs to place is $ \frac{q_{m i_m}}{s_{m i_m}}s_{n i_n}$, where $i_m$ and $i_n$ are the indexes of highest normalized value bids of mobile user $m$ and $n$, respectively. 
	Thus, the payment of winning mobile user $n$ in the pricing scheme is $ g_{n i_n}=\chi_{ni_n}- \frac{q_{m i_m}}{s_{m i_m}}s_{n i_n}$.
	
	\subsection{Properties}
	
	Now, we show that the winner determination algorithm is monotone and the payment determined for a winner mobile user is the difference between the local accuracy based satisfaction and the critical value of its bid. From line 13 of the Algorithm \ref{alg}, it is clear that a mobile user can increase its chance of winning by increasing its bid. Also, a mobile user can increase its chance to win by decreasing the weighted sum of the resources. Therefore, the winner determination algorithm is monotone with respect to mobile user's bids. Moreover, the value of a winning bidder is equals to the minimum value it has to bid to win its bundle, i.e., its critical value. This is done by finding the losing bidder $m$ who would win if bidder $n$ would not participate in the auction. Thus, the proposed mechanism has a monotone allocation algorithm and payment for the winning bidder equals to the difference between the local accuracy based satisfaction and the critical value of its bid. We conclude that proposed mechanism is a truthful mechanism according to Lemma \ref{critical}.
	
	Next, we prove that the proposed auction mechanism is individual rational. For any mobile user $n$ bidding its true value, we consider two possible cases: 
	
	\begin{itemize}
		\item If mobile user $n$ is a winner with its bid $i$th, 
		its payment is 
		\begin{equation*}
		\begin{aligned}
		U_{ni} &= g_{ni} - v_{ni} \\
		&= (\chi_{ni}- \frac{q_{m i_m}}{s_{m i_m}}s_{n i} - v_{ni}\\
		&= \left(\frac{\chi_{ni}-v_{ni}}{s_{ni}}-\frac{q_{m i_m}}{s_{m i_m}}\right)s_{ni} \\
		&= \left(\frac{q_{ni}}{s_{ni}}-\frac{q_{m i_m}}{s_{m i_m}}\right)s_{ni} \geq 0
		\end{aligned}
		\end{equation*}
		where $m$ the losing bidder with the highest normalized valuation if $n$ does not participate in the auction and the last inequality follows from Algorithm \ref{alg}.
		\item If mobile user $n$ is not a winner. Its utility is 0.
	\end{itemize} 
	Therefore, the proposed auction mechanism is individual rational.
	
	Finally, we show that the proposed auction mechanism is computationally efficient.
	We can see that in Algorithm \ref{alg}, the while-loop (lines 12-22) takes at most 
	$N$ times, linear to input. Calculating the payment takes at most $N(N-1)$ times.
	Therefore, the proposed auction mechanism is computationally efficient.
	\vspace{-0.1in}     
	\section{Simulation Results}\label{sec_simulation}
	In this section, we provide some simulation results to evaluate the proposed mechanism. The parameters for the simulation are set the following. The required CPU cycles for performing a data sample $c_n$ is uniformly distributed between $[10, 50]$ cycles/bit. The size of data samples of each mobile user is $s_n=800 \times 10^3$, the maximum tolerance time of a FL task is $T_{max}=[100,500]$. 
	%follows uniform distribution $[5, 8]$ Mbits. 
	The effective switched capacitance in local computation is $\xi = 10^{-26}$. We assume that the noise power spectral density level $N_0$ is $-174$dBm/Hz, the sub-channel bandwidth is $W = 15$ kHz and the channel gain is uniformly distributed between $[-90, -95]$ dB. In addition, the maximum and minimum transmit power of each mobile user is uniformly distributed between $[3, 6]$ mW and between $[1, 2]$ mW, respectively. The maximum and minimum computation capacity is uniformly distributed between $[3, 5]$ GHz and between $[0.1, 0.2]$ GHz, respectively. We also assume that the total number of sub-channels and antennas of the BS are 100 and 100, respectively.
	\begin{figure*}
		\begin{subfigure}[b]{0.5\textwidth}
			%\centering
			\includegraphics[width=0.9\linewidth]{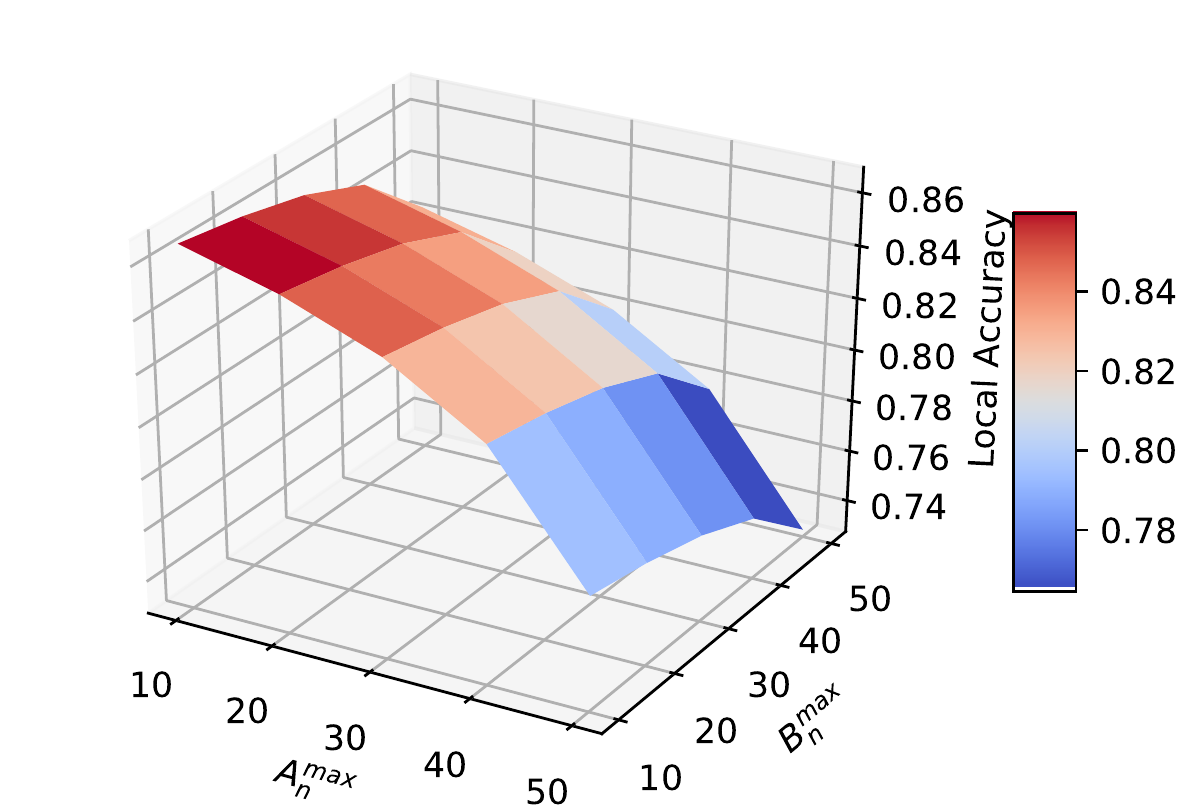}
			\caption{}
			\label{2}
			%\vspace{-1.5em}
		\end{subfigure}
		\begin{subfigure}[b]{0.5\textwidth}
			%\centering
			\includegraphics[width=0.9\linewidth]{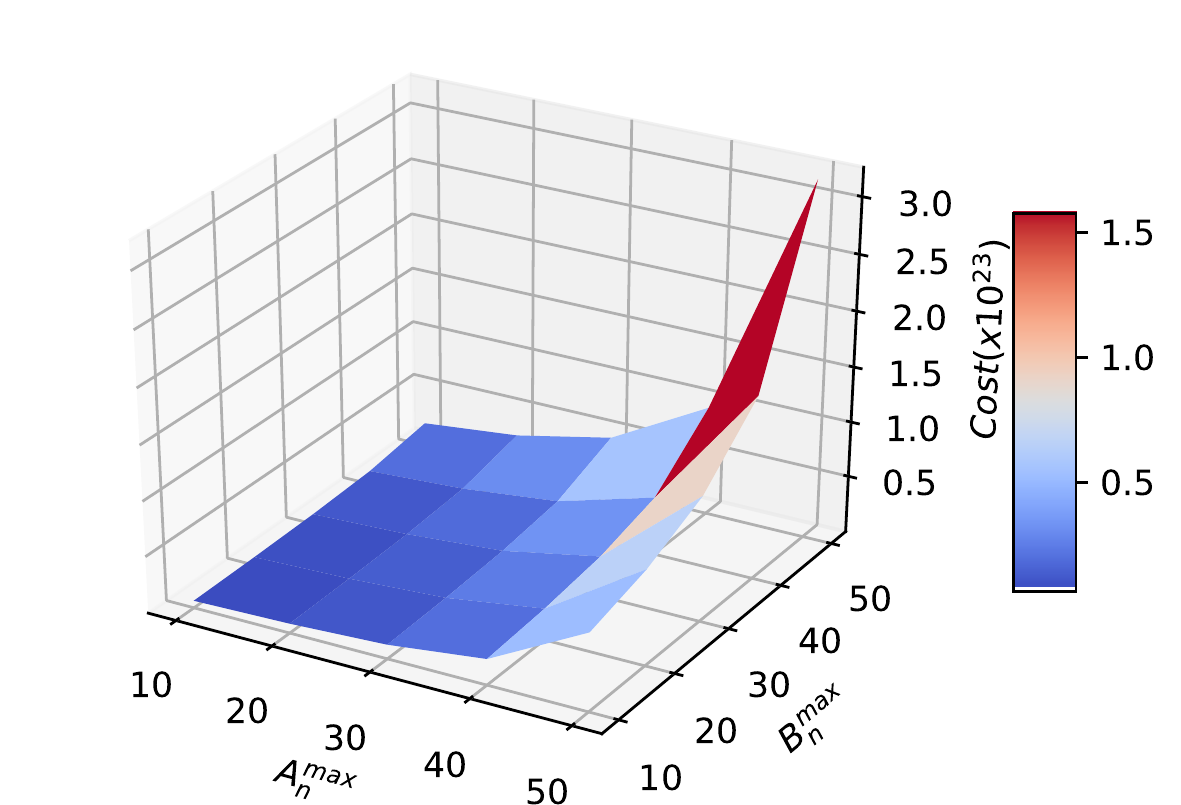}
			\caption{}
			\label{3}
			%	\vspace{-1.5em}
		\end{subfigure}
		\caption{Numerical results a) The changing of the local accuracy when the maximum number sub-channels and antennas in one bid vary, b) The changing of the energy cost when the maximum number sub-channels and antennas in one bid vary.}
	\end{figure*}   
	\begin{figure*}
		\begin{subfigure}[b]{0.5\textwidth}
			%\centering
			\includegraphics[width=0.9\linewidth]{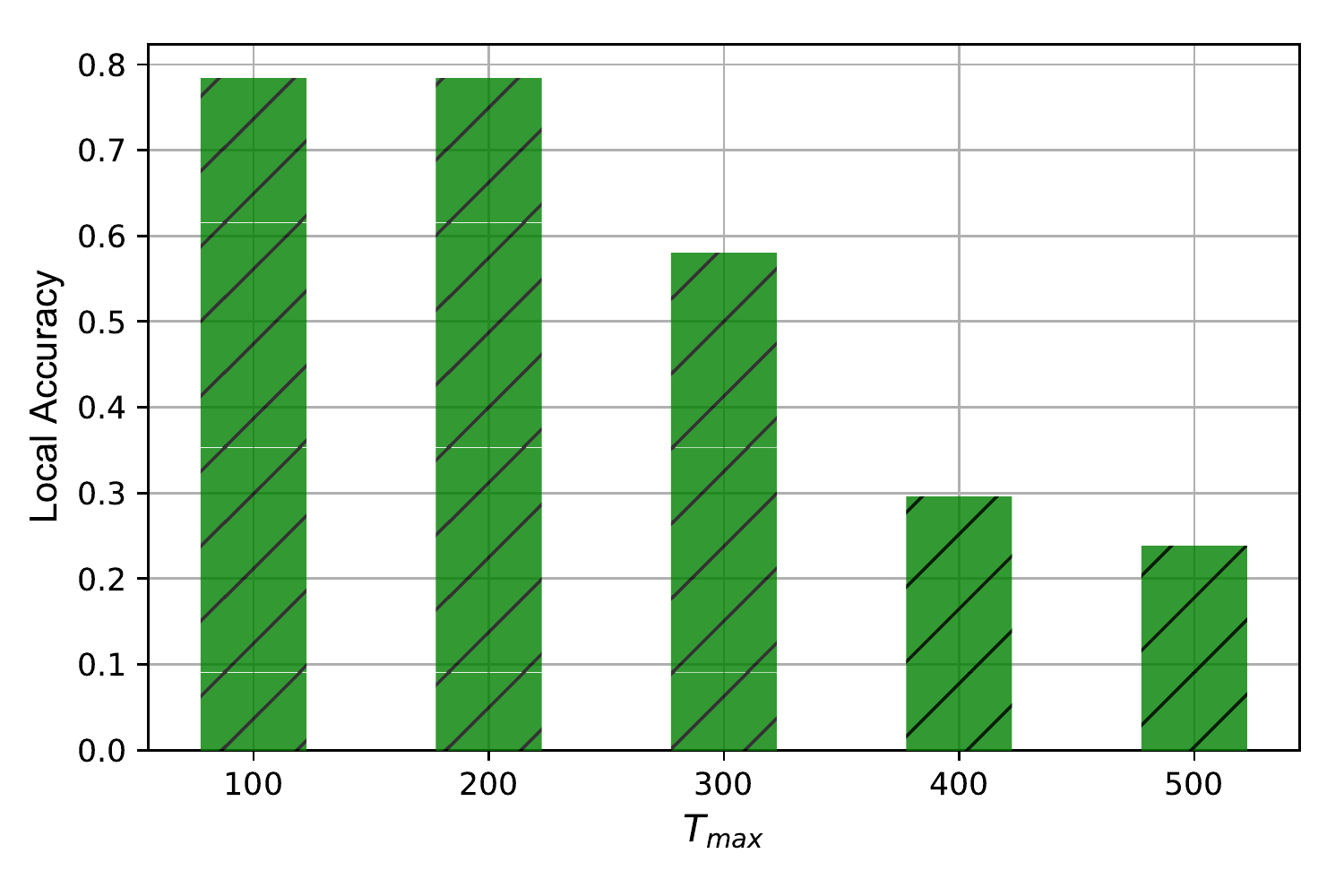}
			\caption{}
			\label{4}
			%\vspace{-1.5em}
		\end{subfigure}
		\begin{subfigure}[b]{0.5\textwidth}
			%\centering
			\includegraphics[width=0.9\linewidth]{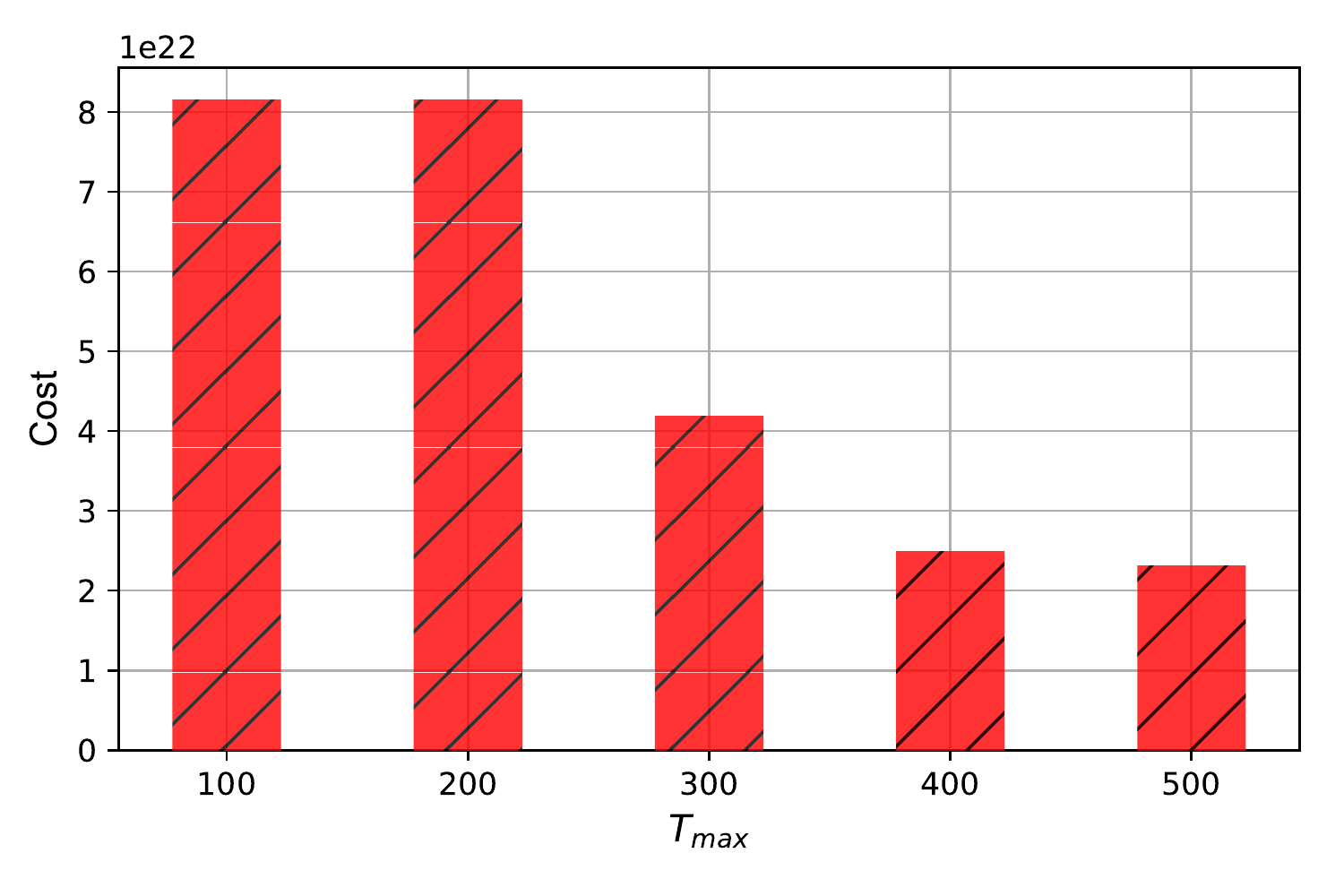}
			\caption{}
			\label{5}
			%	\vspace{-1.5em}
		\end{subfigure}
		\caption{Numerical results a) Local accuracy v.s. $T_{max}$ b) Energy cost v.s. $T_{max}$}
	\end{figure*} 
	\begin{figure*}
		\begin{subfigure}[b]{0.5\textwidth}
			%\centering
			\includegraphics[width=0.9\linewidth]{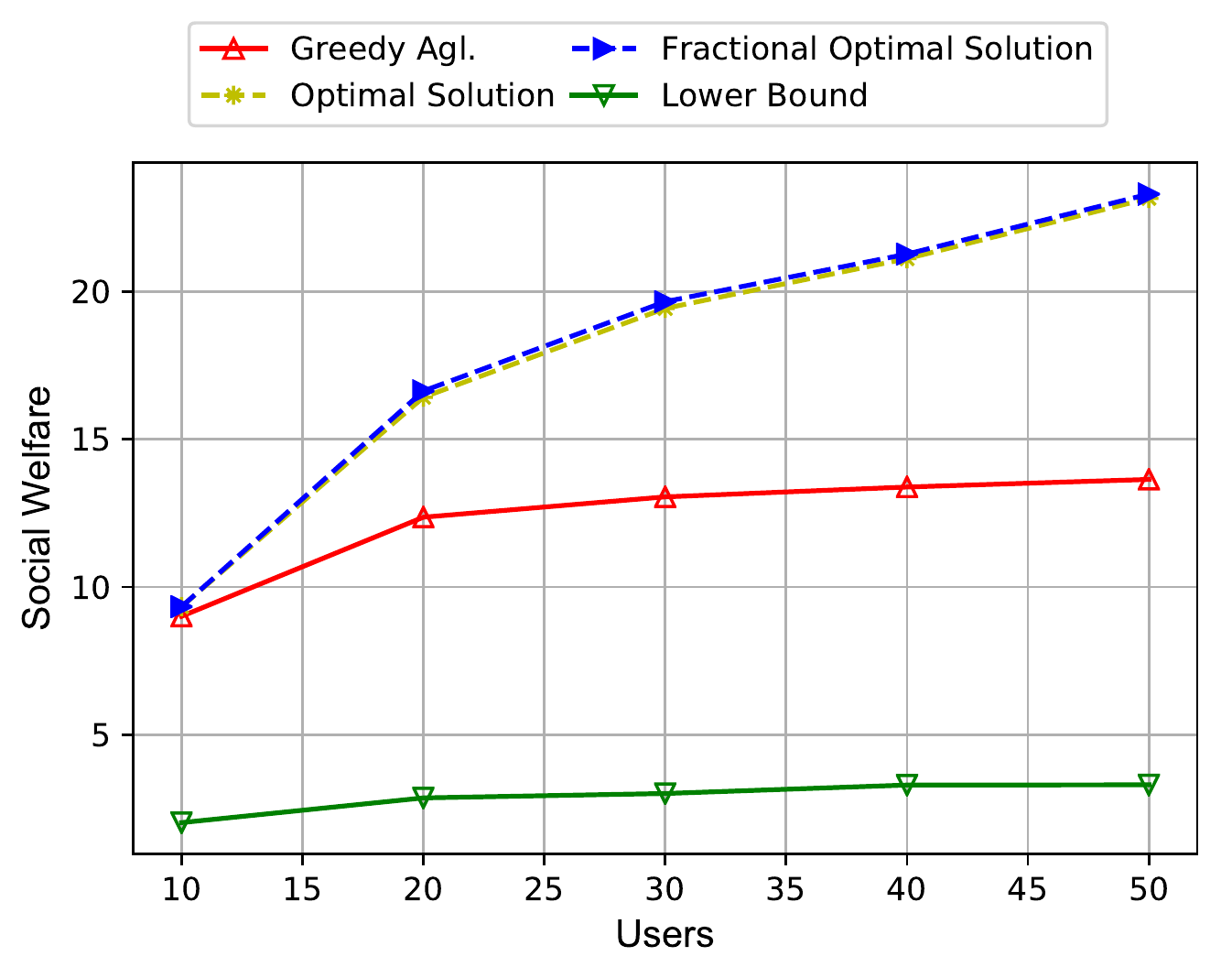}
			\caption{}
			\label{6}
			%\vspace{-1.5em}
		\end{subfigure}
		\begin{subfigure}[b]{0.5\textwidth}
			%\centering
			\includegraphics[width=0.9\linewidth]{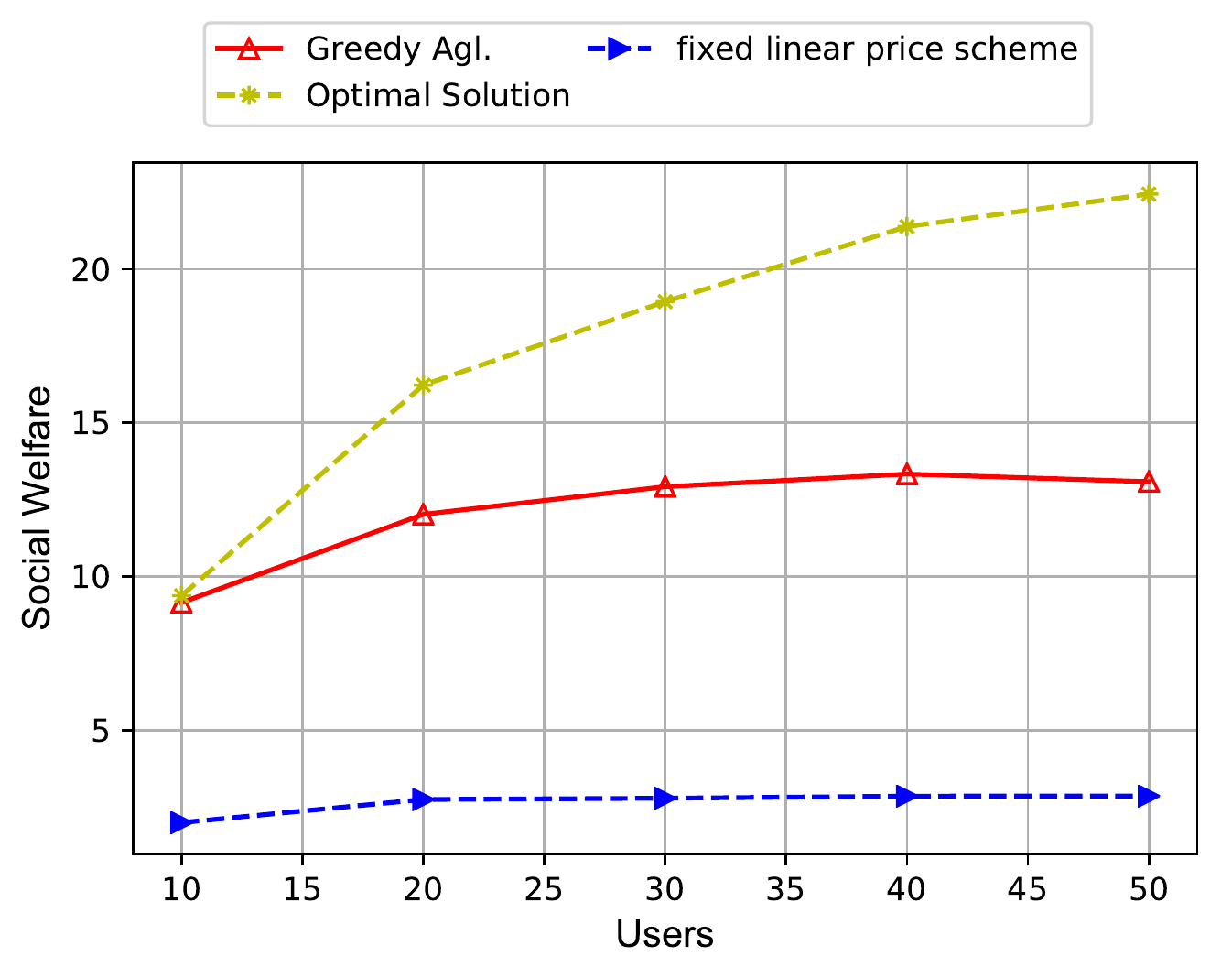}
			\caption{}
			\label{7}
			%	\vspace{-1.5em}
		\end{subfigure}
		\caption{Numerical results for social welfare a) four schemes: Optimal solution, fractional optimal solution, proposed greedy algorithm and lower bound b) three schemes: Optimal solution, proposed greedy algorithm and fixed price scheme.}
	\end{figure*}
	Firstly, we use the iterative Algorithm \ref{alg4} to perform the characteristic of evaluating bids. 
	%Firstly, we perform the characteristic of deciding bids by using the iterative algorithm Alg.\ref{alg4}. 
	The maximum number of sub-channels $B_n^{max}$ and antennas $A_n^{max}$ for mobile user $n$ to request in each bid vary from 10 to 50. 
	Fig.~\ref{2} shows the accuracy level that mobile user $n$ requires to provide decreases when the maximum number of sub-channels $B_n^{max}$ and antennas $A_n^{max}$ increase. 
	However, the decreasing requested local accuracy of mobile user data leads to the increase of global and local rounds to achieve global accuracy. 
	As a result, the cost increases when the number of sub-channels and antennas increases, as shown in Fig.~\ref{3}.

	Fig.~\ref{4} and Fig.~\ref{5} present the cost of one bid of the mobile user and local accuracy, respectively, when the maximum tolerance time $T_{max}$ varies from $100 $ to $500$. When the maximum tolerance time increases, the cost decreases. It is natural because mobile user $n$ can keep low contributing CPU cycle frequency and transmission rate while guaranteeing the delay constraint.
	
	In the following, we evaluate the performance of the proposed auction algorithm. 
	To compare with the proposed algorithm, we use three baselines:
	\begin{itemize}
		\item Optimal Solution: \textbf{P6} is solved optimally.
		\item Fractional Optimal Solution: the linear relaxation of \textbf{P6} is solved optimally.
		\item Fixed Price Scheme \cite{zaman2013combinatorial}: In this scheme, price vector $f=\{f_b,f_a\}$ is the vector that mobile users need to pay for the resource. In this scheme, the mobile users are assumed to be served in a first come, first served basic until the resources are exhausted. The mobile user can get the resource when the valuation of mobile user's bid is at least $F_{ni}=B_{ni}f_b + A_{ni} f_a$ which is the sum of the fixed price of each resource in its bid. We consider three kinds of price vector: linear price ($f_i=f_o\times\eta_i, i=a,b$), sub-linear price vector ($f_i=f_o\times \eta_i^{0.85}, \, i=a,b$) and a super linear price vector ($f_i=f_o\times\eta_i^{1.15}, i=a,b$). Here, we call $f_o$ as the basic price. Unless specified otherwise, we choose $f_o=0.01$.	  
	\end{itemize}
		\begin{figure*}
			\centering
			\begin{subfigure}[b]{0.45\textwidth}				
				\includegraphics[width=0.9\linewidth]{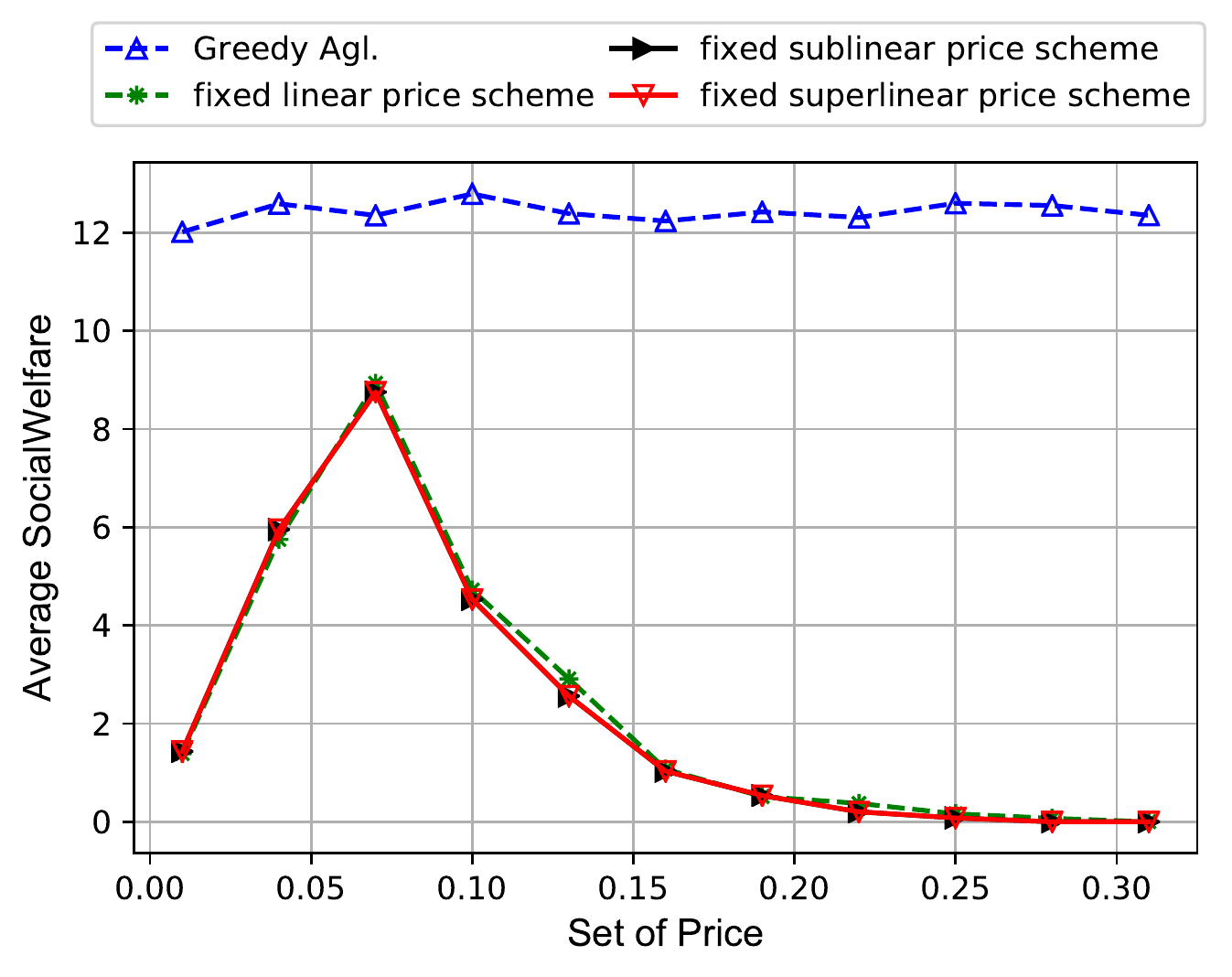}
				\caption{}
				\label{8}
				%\vspace{-1.5em}
			\end{subfigure}
			\begin{subfigure}[b]{0.45\textwidth}
				%\centering
				\includegraphics[width=0.9\linewidth]{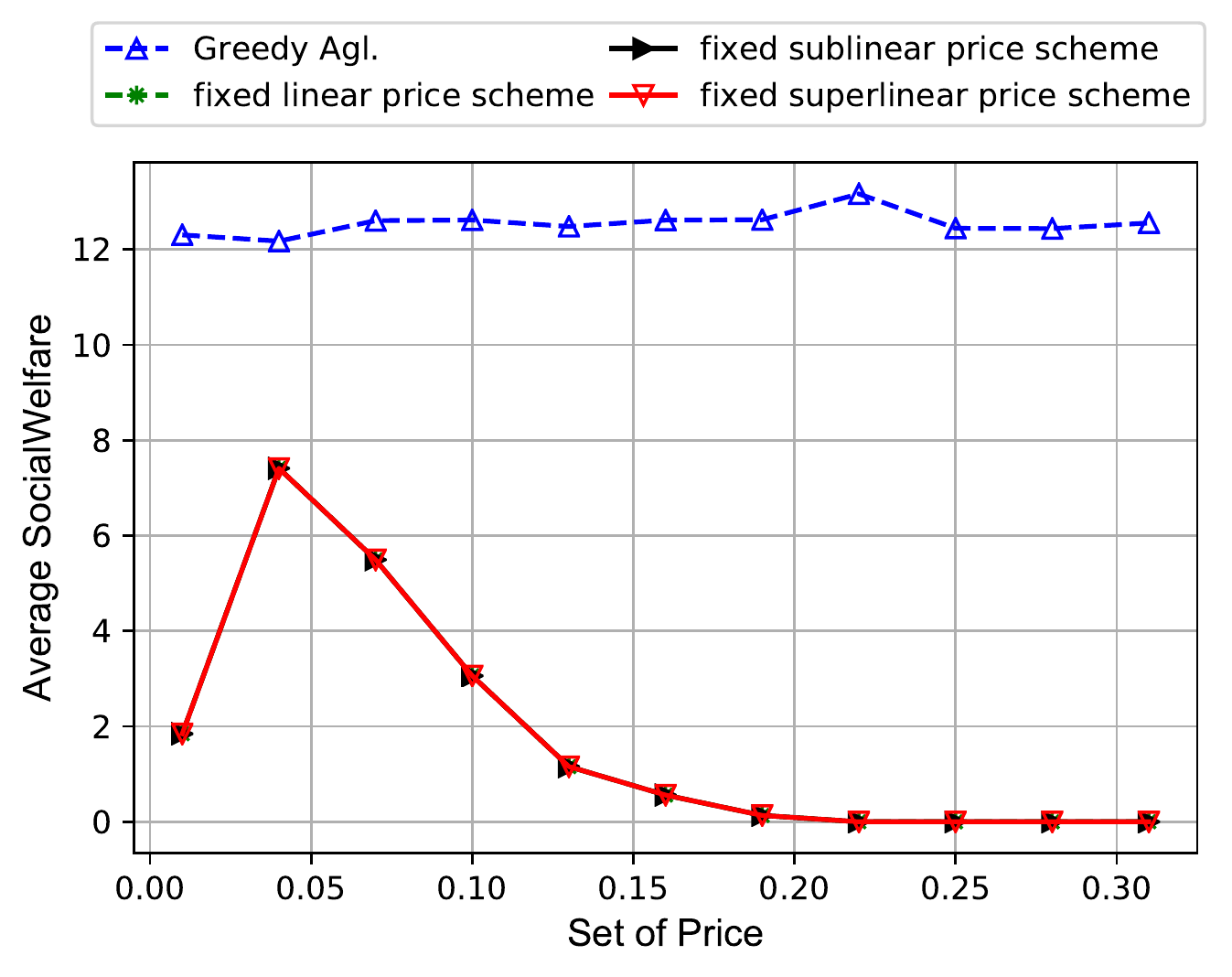}
				\caption{}
				\label{9}
				%	\vspace{-1.5em}
			\end{subfigure}
			\begin{subfigure}[b]{0.45\textwidth}
				\centering
				\includegraphics[width=0.9\linewidth]{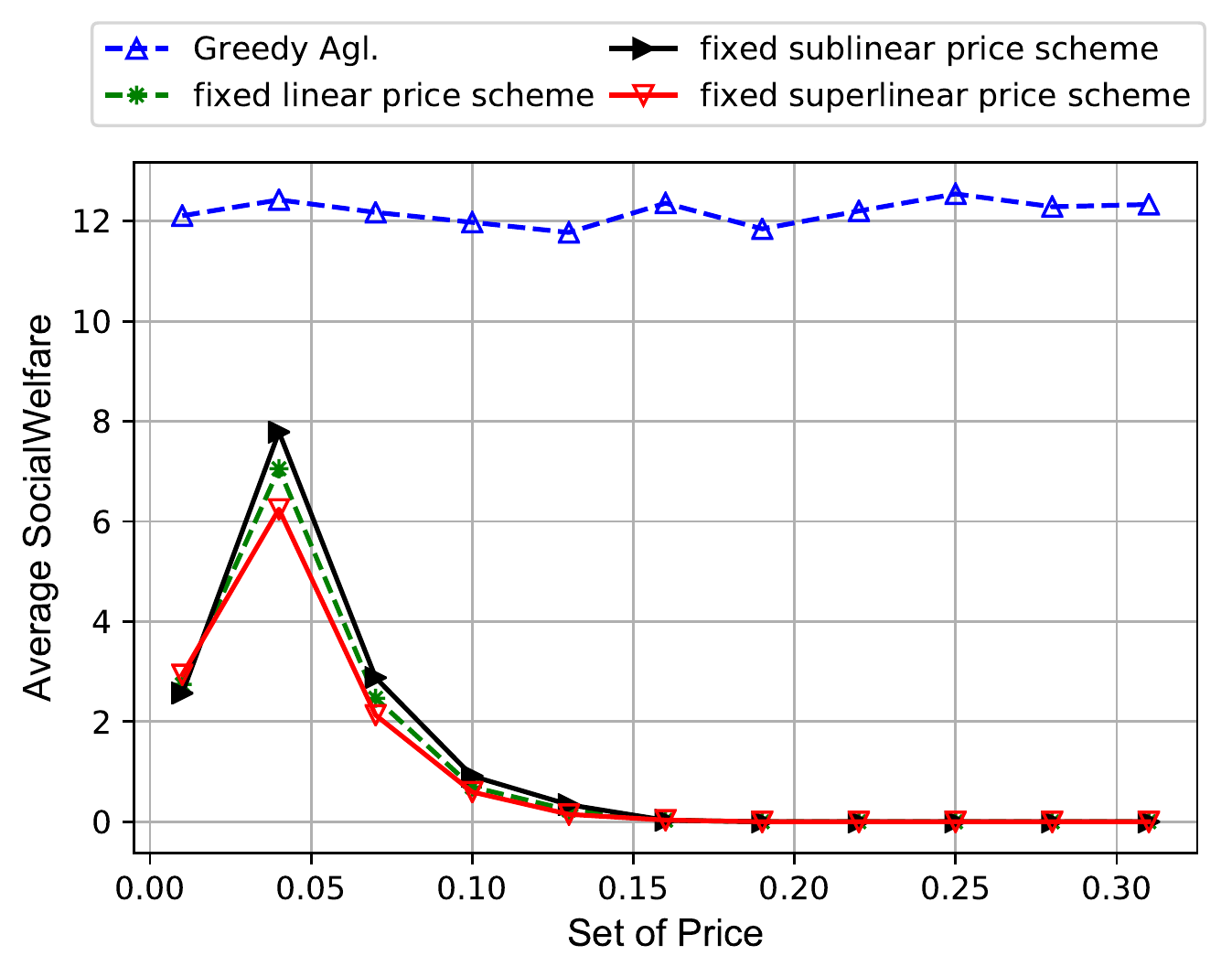}
				\caption{}
				\label{10}
				%\vspace{-1.5em}
			\end{subfigure}
			\caption{Numerical results for social welfare when a) $\eta_a=1, \eta_b=0.5$ , b) $\eta_a=1, \eta_b=1$, c) $\eta_a=1, \eta_b=2$.}
		\end{figure*}      
		\begin{figure*}
			\centering
			\begin{subfigure}[b]{0.45\textwidth}
				%\centering
				\includegraphics[width=0.9\linewidth]{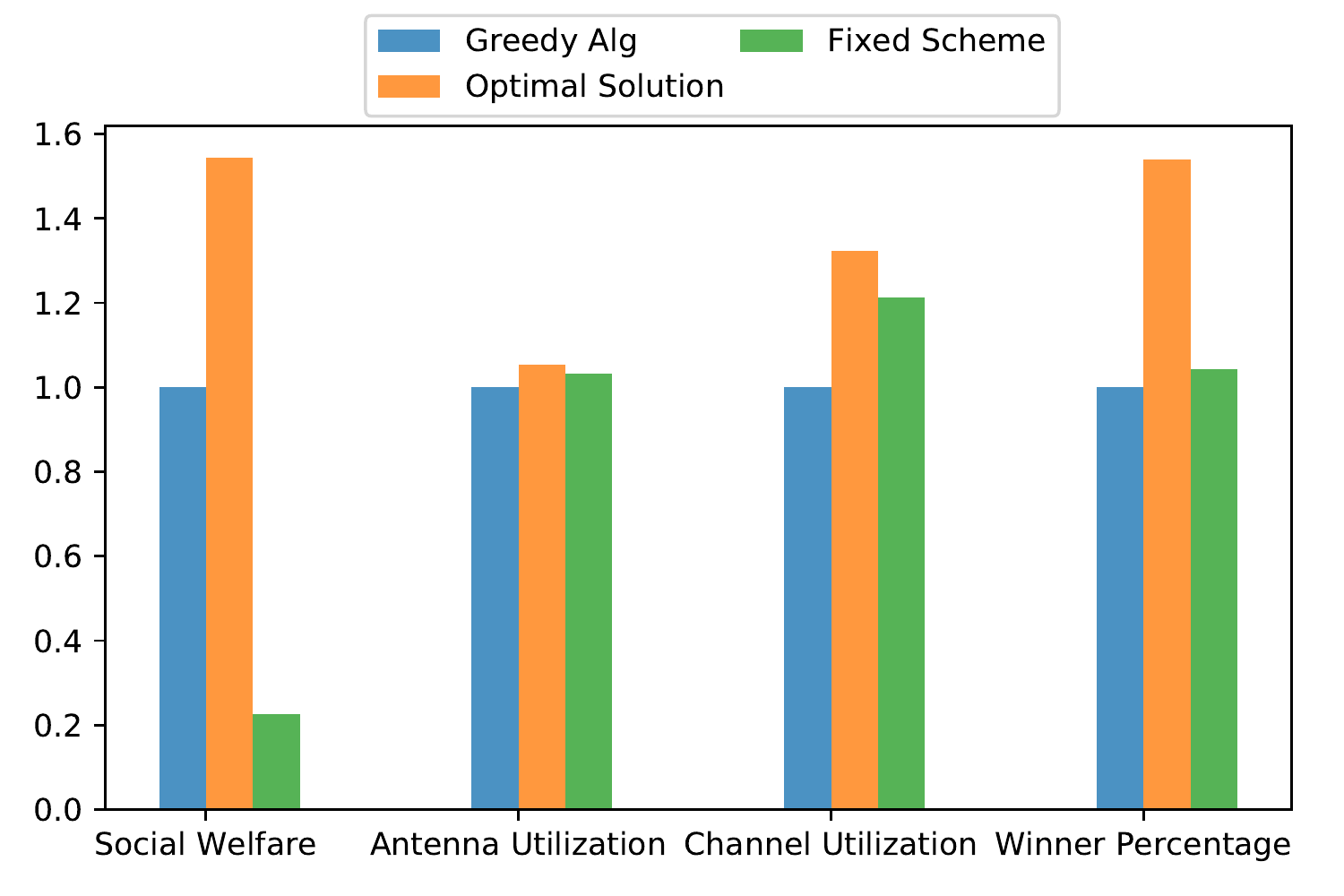}
				\caption{}
				\label{11}
				%\vspace{-1.5em}
			\end{subfigure}
			\begin{subfigure}[b]{0.45\textwidth}
				%\centering
				\includegraphics[width=0.9\linewidth]{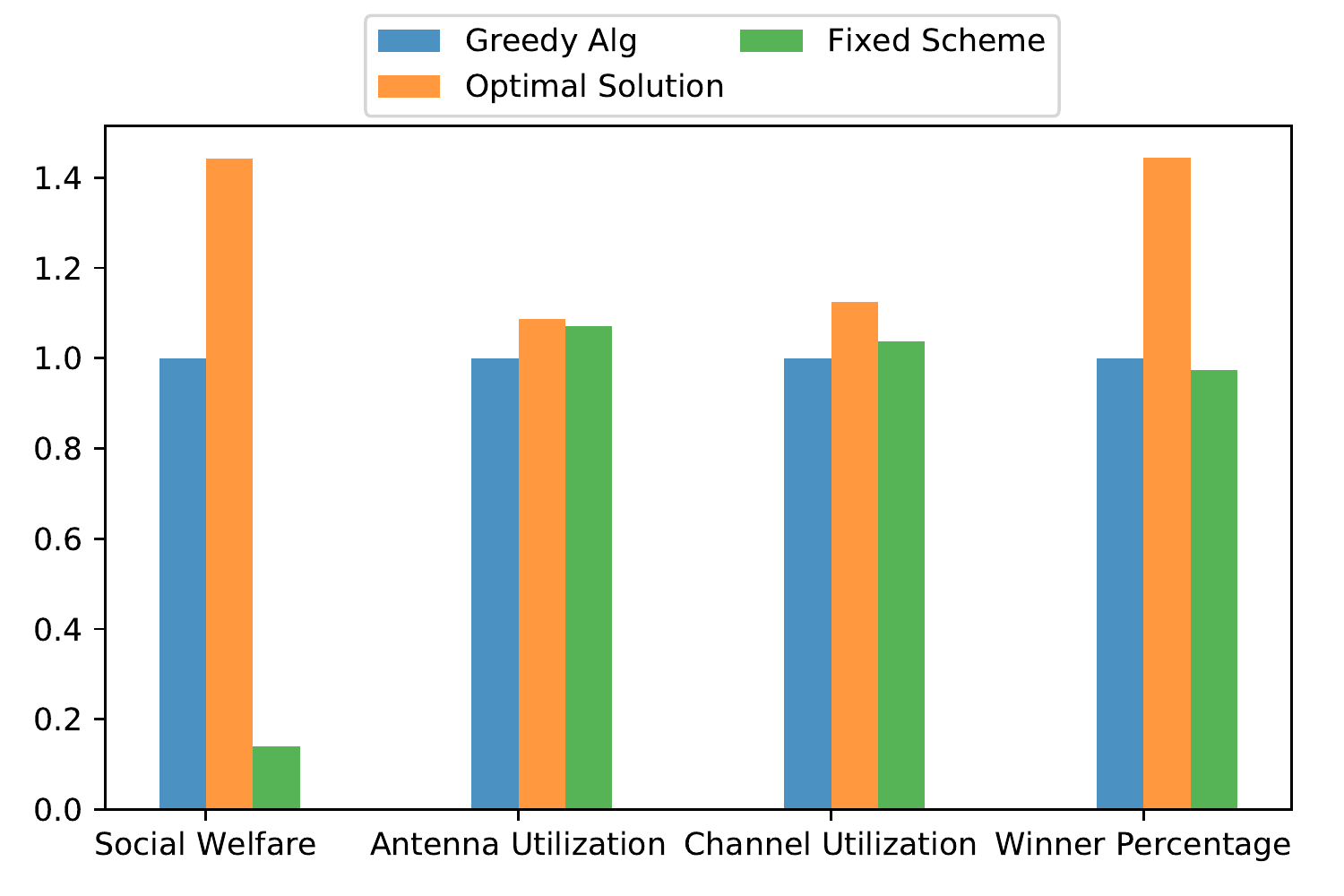}
				\caption{}
				\label{12}
				%	\vspace{-1.5em}
			\end{subfigure}
			\begin{subfigure}[b]{0.45\textwidth}
				\centering
				\includegraphics[width=0.9\linewidth]{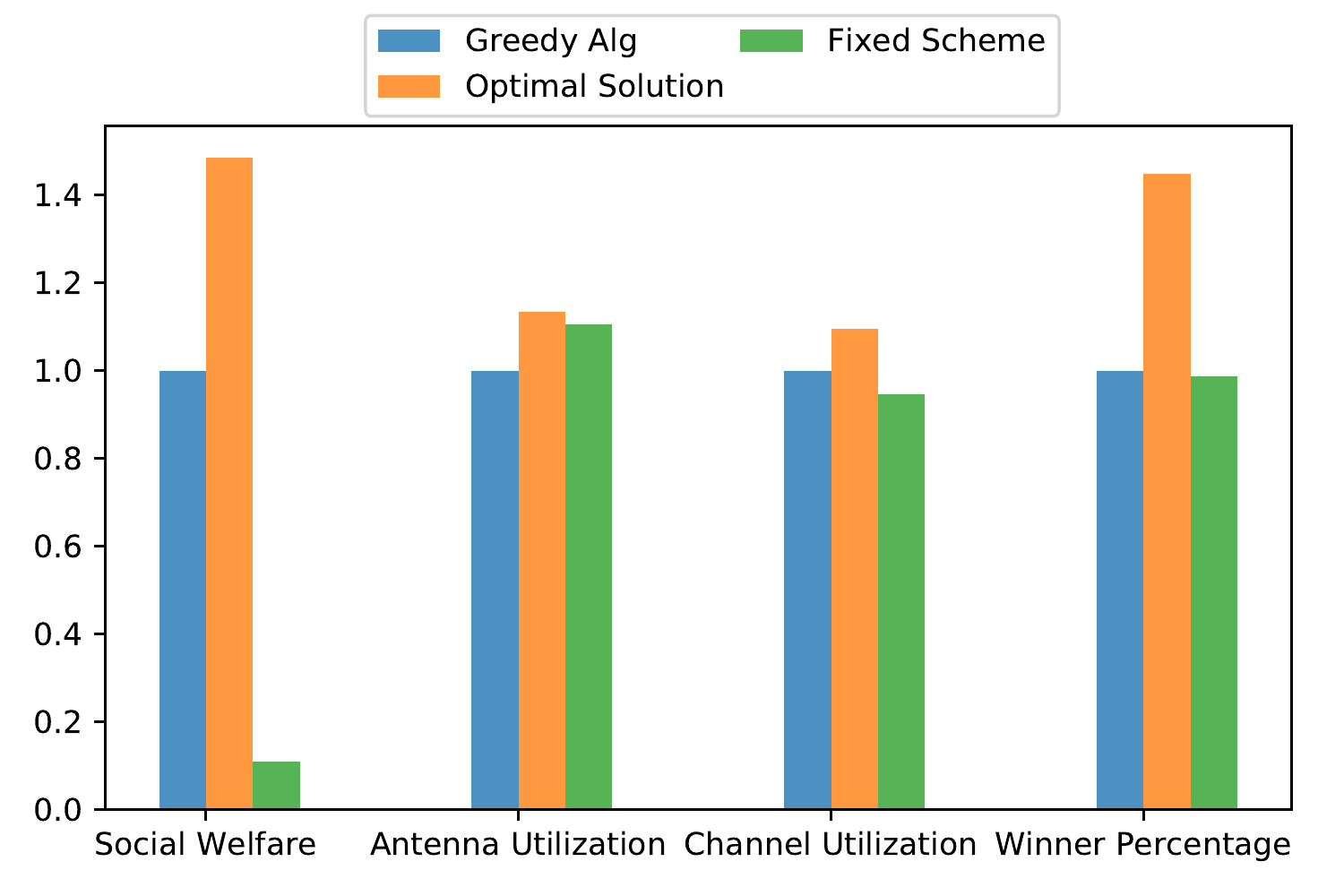}
				\caption{}
				\label{13}
				%\vspace{-1.5em}
			\end{subfigure}
			\caption{Numerical results for normalized ratio when a) $\eta_a=1, \eta_b=0.5$ , b) $\eta_a=1, \eta_b=1$, c) $\eta_a=1, \eta_b=2$.}
		\end{figure*}    
	Fig.~\ref{6} reports the performance of the optimal solution, the fractional optimal solution, the lower bound, and the proposed greedy scheme. The lower bound is determined by the fractional optimal solution divided by gap when the number of mobile users varies from 10 to 50. We note that with the number of mobile users increasing, all schemes produce higher social welfare. This is because there is more chances to choose winning bids with the higher value. Although the social welfare obtained through the proposed greedy scheme is lower than through optimal solution and fractional optimal solution, it much higher than the lower bound.
	
	Fig.~\ref{7} shows the social cost achieved by optimal solution, the proposed greedy scheme and fixed linear scheme when the number of mobile users varies from 10 to 50. We can see that the proposed greedy scheme can provide the much higher social welfare than the fixed linear scheme. 
	
	Since the fixed price scheme heavily depends on the prices of resources, the next experiment helps us to decide whether the fixed-price vector or the performance of the proposed mechanisms is better when we change the basic price $f_o$ between $[0.01, 0.31]$ with the step is 0.03. Fig.~\ref{8}, Fig.~\ref{9} and Fig.~\ref{10} show that the social welfare of fixed price firstly increases and then decreases and equal to 0 when the initial price increases. This is because when the basic price becomes too high, the sum of the price is higher than the valuation of the resources claimed in a bid. Moreover, the social welfare achieved by linear, sublinear and superlinear price schemes are lower than by the proposed greedy scheme. This proves our proposed auction scheme outperforms the fixed price scheme.
	
	In Fig.~\ref{11}, Fig.~\ref{12} and Fig.~\ref{13}, we observe the metrics: social welfare, resource utilization and percentage of three schemes: optimal solution, greedy proposed scheme and fixed price schemes with linear (Fig.~\ref{11}), sublinear (Fig.~\ref{12}), superlinear (Fig.~\ref{13}) fixed price vector. We perform in terms of the ratio with proposed greedy scheme. Among these schemes, the optimal solution is the highest in terms of all metrics. Compared with the proposed scheme, the fixed price can utilize more resources and more mobile users but provides less social welfare. This is due to the fact that the fixed price mechanism heavily depends on the prices of the resources.
	
	\section{Conclusion}\label{sec_conclusion}
	\textcolor{black}{This paper focus on the incentive mechanism design to stimulate mobile users to participate in FL. We formulated the incentive problem between the BS and mobile users in the FL service market as the auction game with the objective of maximizing social welfare. Then, we presented the method for mobile users to decide the bids submitted to the BS so that mobile users can minimize the energy cost. We also proposed the iterative algorithm with low complexity. In addition, we proposed a primal-dual greedy algorithm to tackle the NP-hard winner selection problem. Finally, we showed that the proposed auction mechanism guarantee truthfulness, individual rationality and computation efficiency. Simulation results demonstrated the effectiveness of the proposed mechanism where social welfare obtained by our proposed mechanism is $400 \%$ larger than by the fixed price scheme.}
	\ifCLASSOPTIONcaptionsoff
	\newpage
	\fi

\bibliographystyle{IEEEtran}      
\bibliography{WCNC}	
\end{document}